\documentclass[journal]{IEEEtran}
\usepackage{cite}
\usepackage{amsmath,amssymb,amsfonts}
\usepackage{graphicx}
\usepackage{textcomp}
\usepackage{xcolor}
\usepackage{multirow}
\usepackage{subcaption}
\usepackage{booktabs}
\usepackage{verbatim}
\usepackage{array}
\usepackage{stfloats}
\usepackage{url}
\usepackage{algorithm}
\usepackage{algorithmic}
\usepackage{colortbl}
\usepackage{wrapfig}

\def\BibTeX{{\rm B\kern-.05em{\sc i\kern-.025em b}\kern-.08em
    T\kern-.1667em\lower.7ex\hbox{E}\kern-.125emX}}
\begin{document}
\title{Hierarchy-Aware Neural Subgraph Matching with Enhanced Similarity Measure}
\author{Zhouyang Liu$^*$,
        Ning Liu$^*$,
        Yixin Chen$^\dag$,
        Jiezhong He,
        Menghan Jia,
        Dongsheng Li$^\dag$
\thanks{$*$ means equal contribution. Zhouyang Liu, Yixin Chen, Jiezhong He, Menghan Jia, and Dongsheng Li are with the College of Computer Science and Technology,  National University of Defense Technology, Changsha, Hunan, China (e-mail: \{liuzhouyang20, chenyixin, jiezhonghe, jiamenghan12, dsli\}@nudt.edu.cn). Ning Liu is with the College of Information and Communication, National University of Defense Technology, Wuhan, Hubei, China (e-mail: liuning17a@nudt.edu.cn).}
\thanks{This work is supported in part by National Key Research and Development Program of China (No.
2023YFB4502300) and the National Natural Science Foundation of China under grants (Nos. 62402503,
62025208 and 62421002). \emph{(Corresponding authors: Yixin Chen, Dongsheng Li).}}}

\maketitle

\begin{abstract}
Subgraph matching is challenging as it necessitates time-consuming combinatorial searches. Recent Graph Neural Network (GNN)-based approaches address this issue by employing GNN encoders to extract graph information and hinge distance measures to ensure containment constraints in the embedding space. These methods significantly shorten the response time, making them promising solutions for subgraph retrieval. However, they suffer from scale differences between graph pairs during encoding, as they focus on feature counts but overlook the relative positions of features within node-rooted subtrees, leading to disturbed containment constraints and false predictions. Additionally, their hinge distance measures lack discriminative power for matched graph pairs, hindering ranking applications. We propose NC-Iso, a novel GNN architecture for neural subgraph matching. NC-Iso preserves the relative positions of features by building the hierarchical dependencies between adjacent echelons within node-rooted subtrees, ensuring matched graph pairs maintain consistent hierarchies while complying with containment constraints in feature counts. To enhance the ranking ability for matched pairs, we introduce a novel similarity dominance ratio-enhanced measure, which quantifies the dominance of similarity over dissimilarity between graph pairs. Empirical results on nine datasets validate the effectiveness, generalization ability, scalability, and transferability of NC-Iso while maintaining time efficiency, offering a more discriminative neural subgraph matching solution for subgraph retrieval. Code available at \url{https://github.com/liuzhouyang/NC-Iso}.
\end{abstract}
\section{Introduction}
\IEEEPARstart{S}{ubgraph} matching, or subgraph isomorphism problem, which determines whether a subgraph of a data graph is isomorphic to a query graph, is one of the most fundamental graph operations in real-world applications such as subgraph retrieval and social network analysis. Conventional approaches usually formulate subgraph matching as a combinatorial search task. They determine subgraph matching by finding exact bijective node projections between query graphs and subgraphs within data graphs \cite{Ullmann,  VF2, SAGA, QuickSI, Turboiso, RI, CFL, VF3, VF2++, DP-iso}, which suffer from exponential time complexity. 

To address this problem and enable fast responses for tasks such as subgraph retrieval, neural subgraph matching (NSM) methods have emerged as a promising solution. Instead of finding bijective node projections, NSM methods predict subgraph matching to prune unmatchable graphs as much as possible to reduce the search space. Then, subgraph matching algorithms can be applied to further enumerate the matches if needed. To this end, they employ Graph Neural Network (GNN) encoders to learn graph representations in the encoding stage. Unlike conventional algorithms preprocess graph pairs in an on-the-fly manner, these learned representations of NSM methods can be reused, further saving computational resources and facilitating querying operations. Subsequently, NSM methods determine subgraph matching through hinge distance measures in the scoring stage. 

\begin{figure}[t]
\centering
\includegraphics[width=0.7\linewidth]{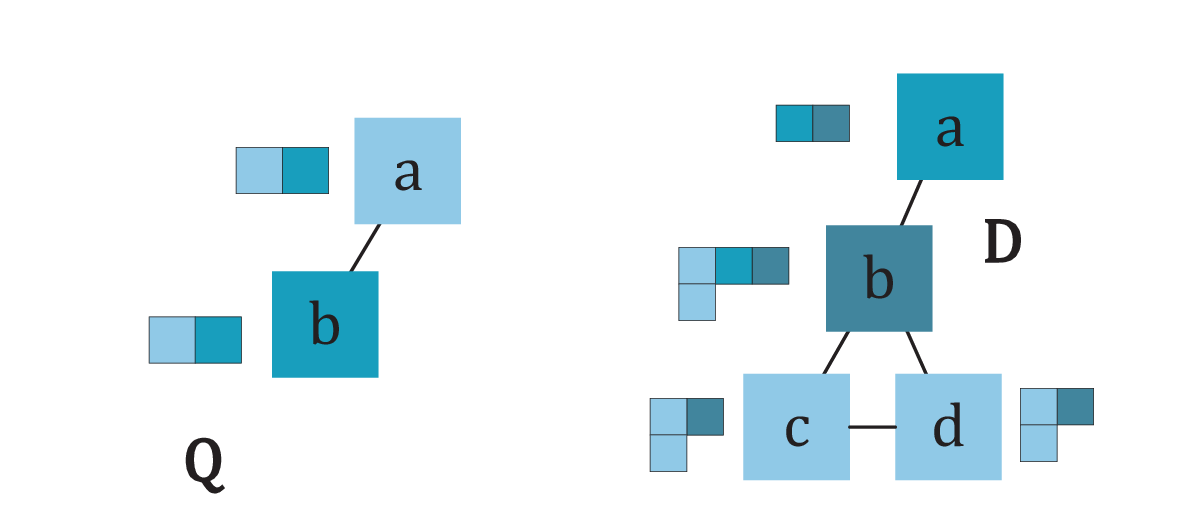}
\caption{The $2$-hop label counts of nodes within $Q$ and $D$, represented by black-bordered squares. Due to scale differences, node $b$ in $D$'s $k$-hop label count exceeds that of nodes in $Q$, leading to a false positive match despite no structural alignment. Additionally, the additive nature of feature counts across nodes results in larger sums in data graphs, undermining coarse-grained graph-level containment constraints and exacerbating matching errors.} \label{fig:hier_motive}
\vspace{-0.5cm}
\end{figure}

These methods relax subgraph isomorphism to detect compliance with containment constraints, a necessary condition for subgraph isomorphism, between the representations as an inductive bias. Specifically, NeuroMatch \cite{Ying2020NeuralSM} and IsoNet \cite{Roy2022InterpretableNS} ensure that the representations of data graphs/edges include these of candidate query graphs/edges in all dimensions. D2Match \cite{liu2023d2match} requires containment in the neighborhoods of matched data-query node pairs. These constraints reduce the NP-complete subgraph matching problem \cite{Hjorth2005TJS} to a polynomial-time approximation \cite{liu2023d2match}, leading to increased efficiency. 
Hinge distance measures further serve to detect the violations of these constraints, quickly filtering out unmatchable pairs. In particular, fine-grained node/edge comparison-based methods evaluate the overall violation of a node/edge alignment learned with graph matching techniques like the Gumbel-Sinkhorn network \cite{Roy2022InterpretableNS} or perfect bipartite matching \cite{liu2023d2match}. However, these graph matching techniques, which force a one-to-one correspondence, may struggle with unmatchable graph pairs that only have partial matches in the subgraph isomorphism setting. Additionally, these methods suffer from quadratic complexity in node/edge counts and may not efficiently handle graphs comprising hundreds of nodes/edges. In contrast, coarse-grained NSM approaches, which directly assess the violation at the graph level \cite{Ying2020NeuralSM}, offer a more favorable solution for subgraph retrieval, as they exhibit efficiency in both training and inference.

Despite the initial success, existing coarse-grained methods still face two main challenges: (\romannumeral1) \emph{They may struggle to handle scale differences in the encoding stage.} 
Prior coarse-grained approaches impose containment constraints on representations generated by GNN encoders. They focus on the counts of node-specific features, such as node labels, within node-rooted subtrees but overlook how features are organized. As a result, nodes with large subtrees may inadvertently contain unmatchable nodes with smaller subtrees in terms of feature counts, disturbing the containment constraints and potentially leading to false prediction. We illustrate the impact of scale differences on subgraph matching in Fig. \ref{fig:hier_motive} for better understanding.
(\romannumeral2) \emph{Their hinge distance measures are less discriminative for matched pairs.} Under current measures, all matched pairs receive a zero distance. Although satisfying containment compliance, it impedes the ability to rank the matched pairs, which is crucial for retrieval and recommendation systems \cite{rank4, rank5} to find the best matches, and can help applications such as drug discovery \cite{rank3} and social network analysis \cite{rank1, rank2} to prioritize which matches to investigate further. In drug discovery, for example, although multiple subgraphs may satisfy the subgraph isomorphism condition with a query graph, additional structural differences can lead to significant variations in chemical properties or biological activity. Ranking is therefore essential to identify the most promising matches.

Here, we present \textbf{NC-Iso}, \textbf{N}eural \textbf{C}ontainment-based Subgraph \textbf{Iso}morphism Predictor, a simple yet effective neural architecture for neural subgraph matching. 
(\romannumeral1) 
To mitigate the influence of scale differences in the encoding stage, NC-Iso proposes preserving the organization of features within node-rooted subtrees, i.e., their relative positions. As the relative positions between nodes are revealed by edges, and edges within subtrees link adjacent echelons, NC-Iso simplifies the preservation of relative positions between nodes to build the hierarchical dependencies between adjacent echelons within node-rooted subtrees, ensuring that matched graph pairs comply with containment constraints in feature counts while maintaining consistent hierarchies. 
(\romannumeral2) To tackle the limitations of hinge distance measures employed in prior work, we draw inspiration from GIoU loss \cite{giou} and propose a novel similarity measure. This measure normalizes the hinge distance as a compliance score of the containment constraint to handle extreme values that may distort the distances. Moreover, it quantifies the extent of similarity and dissimilarity between graph pairs as the similarity dominance ratio (SDR), empowering NC-Iso the ability to rank matched pairs. 
The main contributions of this paper are three-fold: 
\begin{itemize}
    \item We propose NC-Iso, a hierarchy-aware neural architecture that builds the hierarchical dependencies between each echelon within node-rooted subtrees to reduce the influence of scale differences and ensure that matched graph pairs maintain consistent hierarchies. 
    \item We propose a novel similarity measure that normalizes the hinge distance as a compliance score of containment constraint and quantifies the extent of similarity and dissimilarity between graph pairs, providing a more effective and flexible evaluation.
    \item We conduct extensive experiments on nine benchmark datasets for subgraph matching tasks. Comparisons between eight neural and seven conventional baselines validate the effectiveness, generalization ability, and transferability of our proposed method.
\end{itemize}
\section{Related work} \label{sec:Related}

\textbf{Conventional Subgraph Matching Algorithms}. These algorithms can be commonly divided into two categories: exact methods and approximate ones. Approximate approaches allow for mismatch tolerance on the node level \cite{SAGA, Ness, NeMa} while exact algorithms do not \cite{VF2, QuickSI, RI, Turboiso, CFL, VF3, DP-iso, VF2++}. Following \cite{Ullmann}, these algorithms solve subgraph matching by identifying all occurrences of query graphs within data graphs, which requires exploring all possible combinations of subgraphs, making them computationally expensive and less scalable for larger query graphs. Generally speaking, designing such an algorithm is a game of balancing the effectiveness of pruning strategies and computational expense, and the generalizability of such an algorithm is restricted by its heuristic strategies. 

\textbf{Neural Subgraph Matching}.
Recently, researchers have proposed GNN-based methods to accelerate subgraph matching \cite{Ying2020NeuralSM, Roy2022InterpretableNS, liu2023d2match}. 
NeuroMatch \cite{Ying2020NeuralSM} employs order embedding \cite{Vendrov2016OrderEmbeddingsOI} to model containment constraints between matched graphs. Conversely, IsoNet \cite{Roy2022InterpretableNS} and D2Match \cite{liu2023d2match} check compliance with containment constraints at the edge or node level, utilizing techniques like the Gumbel-Sinkhorn network or perfect bipartite matching to estimate the optimal total violation. Despite enhanced efficiency, these models fail to handle partial matches or struggle with scale differences between graph pairs, potentially compromising performance. Additionally, they often rely on hinge distance measures to detect containment compliance, which can falter with extreme distance values and fail to discern matched pairs, limiting applications requiring ranking.
\section{Preliminaries}
\textbf{Notation.} We consider undirected, connected, node-labeled graphs. Let $G = (\mathcal{V}_G, \mathcal{A}_G, \Gamma_G, \mathbf{X}_G)$ be a graph with vertex collection $\mathcal{V}_G$ whose cardinality is $|\mathcal{V}_G|$, adjacency matrix $\mathcal{A}_G \in\{0,1\}^{|\mathcal{V}_G|\times |\mathcal{V}_G|}$, label table $\Gamma_G$, and feature set $\mathbf{X}_G$, which abstractly represents node-specific information such as labels, degree, triangle counts, or other descriptors. In this work, we initialize features using node labels, but the formulation can be extended to other forms of features without loss of generality. In this context, each node $v$ is associated with a label $y_v\in\Gamma_G$. Thus, a graph can be represented by a multiset: $\mathcal{M}_G=(y, C_\mathcal{M})$, drawn from $\Gamma_{G}$ and represented by a function $C_{\mathcal{M}}:\Gamma_G\rightarrow\mathbb{N}$, indicating the number of occurrences of the element $y\in\Gamma_G$ in $\mathcal{M}_G$. Let $\mathcal{M}_1$ be a subset of $\mathcal{M}$ such that $\forall y \in \Gamma_{\mathcal{M}}, C_{\mathcal{M}}(y) \geq C_{\mathcal{M}_1}(y)$.
Given a data graph $D$, $D^\prime$ is a subgraph of $D$, such that $\mathcal{V}_{D^\prime} \subseteq \mathcal{V}_D$, $\Gamma_{D^\prime} \subseteq \Gamma_D$ and $\mathcal{A}_{D^\prime}$ is a submatrix of $\mathcal{A}_{D}$. $\mathcal{N}^k(v)$ represents the $k$-hop neighborhood of $v$, where $\mathcal{N}^0(v)=\{v\}$.

\textbf{Definition} \emph{(Subgraph Isomorphism)}\textbf{.} $Q$ is isomorphic to $D^\prime$, a subgraph of $D$, if there exists a \emph{bijective} function $P: \mathcal{V}_Q \mapsto \mathcal{V}_{D^\prime}$, such that (1)  $\forall v \in \mathcal{V}_Q$, $y_v=y_{P(v)}$ if exists, and (2) $\forall \mathcal{A}_Q(v,v^\prime) = 1, \mathcal{A}_{D^\prime}(P(v),P(v^\prime)) = 1$. $P$ is called a \emph{solution}, $(Q,D)$ is a \emph{subgraph isomorphism}. We further denote subgraph isomorphism pair $(Q, D)$ as $Q \subseteq D$.

\textbf{Observation 1} \emph{(Containment Property). For any $(Q\subseteq D)$ pair, the solution $P$ can be represented by a node permutation matrix $\Pi\in\{0,1\}^{|\mathcal{V}_Q|\times|\mathcal{V}_D|}$, where $\Pi_{ij} = 1$ indicates a match between node $i$ and node $j$. After permutation, $\forall (\mathcal{A}_Q)_{i,j}=1$, there exists $(\Pi\mathcal{A}_D\Pi^\top)_{i,j}=1$, $\mathcal{A}_Q$ becomes a submatrix of $\Pi\mathcal{A}_D\Pi^\top$, which can be expressed as follows:
\begin{equation}\label{eq:1}
\begin{cases}
    (\Pi\mathcal{A}_{D}\Pi^\top - \mathcal{A}_Q)_{i,j} \in \{1, 0\}\\
    \sum_{i,j}(\Pi\mathcal{A}_{D}\Pi^\top - \mathcal{A}_Q)_{i,j} \geq 0
\end{cases}
\end{equation}
The upper equation represents the fine-grained containment of $(Q\subseteq D)$ at the node/edge level. The lower inequality means the sum of the entries is non-negative, indicating a coarse-grained containment at the graph level. This property can be effortlessly extended to subgraph levels, where the $k$-neighborhoods of data nodes/edges contain the counterparts of their corresponding query nodes/edges.} 

\textbf{GNN-based Containment Constraint}. Given the input graph as $\mathcal{M}=\{\mathbf{X}, C_\mathcal{M}\}$, a multiset of features. The containment constraints focus on the counts of features. The Message-passing Graph Neural Networks (MPNNs) operate within the node-rooted subtrees and aggregate over the multisets of neighboring node features to generate representations for the root nodes. For simplicity, we refer to MPNN as GNN in the following. For any matched node pair $(v, P(v))$, the $k$-hop neighborhood of $v$ is isomorphic to a subgraph of $P(v)$ centered $k$-hop neighborhood. As a result, the $k$-order subtree of $P(v)$ contains $v$'s during aggregation. This implies the aggregated information for $P(v)$ inherently contains $v$'s. This connection motivates the use of GNNs in subgraph matching. 
The encoding process for node $v$ at $j$-th layer can be represented as follows. 
\begin{align}
\mathbf{M}_{v}^{j} = \phi^{j}(\operatorname{Combine}^{j}(\mathbf{M}^{j-1}_{v},\operatorname{Aggr}^{j}(\mathbf{M}^{j-1}_{v^{\prime}}: v^\prime \in {\mathcal{N}(v)})))\label{eq:comb}
\end{align}
Where $\mathbf{M}_v^j$ is the representation of $j$-order multiset of $v$, $\mathcal{N}(v)$ is the collection of $v$'s direct neighbors. $\operatorname{Aggr}^j(\cdot)$ aggregates information from $v$'s neighborhood. $\operatorname{Combine}^j(\cdot)$ merges $v$'s representation from the previous layer with aggregated neighborhood information. $\phi^j(\cdot)$ is a learned hash function. 
Since graphs are unordered, permutation-invariant aggregation and combination functions such as $\operatorname{Sum}, \operatorname{Mean}, \operatorname{Max}$, are common choices. To reflect containment constraints, for $v\in Q$ to match $u \in D$, the representation $\mathbf{M}_u$ should contain $\mathbf{M}_v$ at each dimension. For $Q \subseteq D$, each node in $Q$ should be contained by at least one node in $D$, thus $\mathbf{M}_D$ should contain $\mathbf{M}_Q$ at each dimension.
\begin{figure*}
\centering
\begin{minipage}{0.35\linewidth}
\includegraphics[width=1.0\linewidth]{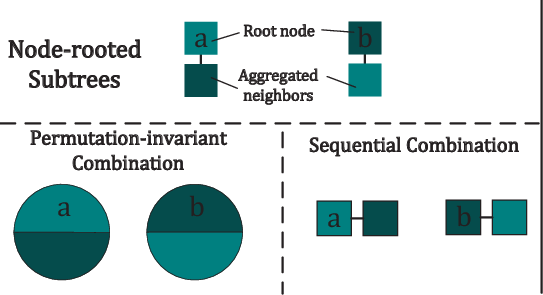}\label{fig:reception}
\end{minipage}
\begin{minipage}{0.35\linewidth}\includegraphics[width=1.0\linewidth]{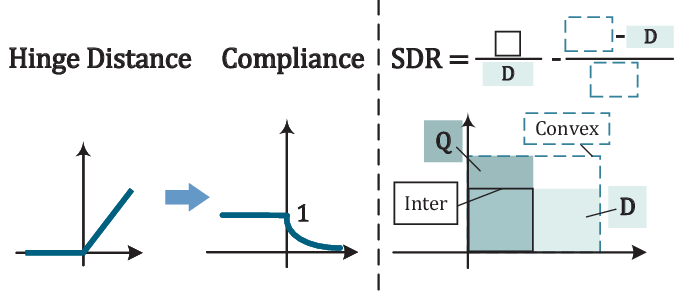}\label{fig:measure}
\end{minipage}
\caption{(Left) GNNs that use permutation-invariant combination function may struggle to distinguish $S_a$ and $S_b$. In contrast, the sequential combination introduces hierarchy awareness, rendering $S_a$ and $S_b$ distinguishable. (Right) Our proposed measure normalizes the hinge distance and considers the intersection and normalized difference between compared pairs.}\label{fig:key}
\vspace{-0.5cm}
\end{figure*}

\section{Analyzing Neural Subgraph Matching}

\textbf{NSM Problem Formulation}. Given graph pairs as input, neural subgraph matching approaches generally have two stages: (1) Encoding stage: project the feature sets $\mathbf{X}$ of graphs into an embedding space $\Phi: \mathbf{X} \mapsto \mathbb{E}^{|\mathcal{V}|\times d}$; (2) Scoring stage: based on a distance/similarity measure function $\Psi$, fine-grained node/edge comparison-based methods 
evaluate the total cost of a learned alignment, whereas coarse-grained graph comparison-based approaches assess the cost at the graph level. 

\subsection{The Influence of Scale Differences in The Encoding Stage}
Prior NSM methods focus on imposing containment constraints on representations generated by GNN encoders, which primarily consider the counts of features rather than their organization, i.e., their relative positions. As a result, these methods suffer from scale differences between graph pairs, referring to variations in the size and complexity of node-centered subtrees (ego-graphs) across graphs. These scale differences can lead to larger graphs containing smaller ones based solely on feature counts, even in the absence of structural alignment, thus invalidating containment constraints. Since GNNs operate within node-rooted subtrees, we extend the definition of subgraph isomorphism to the subtree level to highlight the significance of relative positions between features or nodes within these subtrees.

\textbf{Subtree-level Subgraph Isomorphism.} \emph{For any matched $(v,u)$ pair, $S_v \subseteq S_u$, where $S$ is the node-rooted subtrees, there exists a solution $P$, such that (1) all nodes $n_v \in S_v$ can find a distinct correspondence $P(n_v)\in S_u$, (2) for all $\mathcal{A}_{S_v}(n_v,n_{v^\prime}) = 1$, there exists $\mathcal{A}_{S_u}(P(n_v),P(n_{v^\prime})) = 1$.}

The edge connections reveal the relative positions between nodes. The second requirement above highlights the importance of relative positions and can be extended to the $k$-hop relative position consistency between matched graph pairs, where for any $k$, $\forall \mathcal{A}_{S_v}^{k}(n_v,n_{v^\prime}) = 1$, there exists $\mathcal{A}_{S_u}^{k}(P(n_v),P(n_{v^\prime})) = 1$, here $\mathcal{A}^{k}$ is the $k$-th power of an adjacency matrix.
Within node-rooted subtrees, the edges interconnect adjacent echelons of subtrees. Thus, relative position preservation can be reduced to maintaining the hierarchical dependencies between each adjacent echelon. In contrast, GNNs establish connections between the rooted nodes and their aggregated neighborhoods at the combination step. This similarity guides our focus toward the combination function within GNNs as the crucial element for preserving hierarchical dependencies. Subsequently, we observe that commonly used permutation-invariant combination functions in prior work \cite{Ying2020NeuralSM, liu2023d2match} may cause the loss of hierarchical dependencies. These functions treat the root nodes and aggregated neighbors as unordered multisets of features, resulting in what we identify as subtree hierarchy insensitivity.

\textbf{Subtree Hierarchy Insensitivity.} \emph{Given any root node representation $\mathcal{M}_r$, and its aggregated neighborhood information $\mathcal{M}_n$, along with a permutation-invariant combination function $\operatorname{Combine}(\cdot, \cdot)$:}\\
\emph{1) In the absence of any bias term, $\operatorname{Combine}(\cdot, \cdot)$ satisfies: \begin{equation}
            \operatorname{Combine}(\mathbf{M}_r,\mathbf{M}_n) = \operatorname{Combine}(\mathbf{M}_n,\mathbf{M}_r)\notag
        \end{equation}
        GNNs employing such a $\operatorname{Combine}(\cdot)$ fail to differentiate the root node from its neighbors.}\\
\emph{2) When $\operatorname{Combine}(\cdot, \cdot)$ includes a bias term $\xi$ on root nodes or neighborhoods, it encodes hierarchy-weighted multisets but does not inherently capture the hierarchical dependencies within subtrees.}

This issue is illustrated in Fig. \ref{fig:key} (Left). 
As a result, containment constraints in prior work are subtree-augmented multiset containment, which only focuses on the feature counts within subtrees, suffering from scale differences.
This limitation hinders the distinction between subtrees that, while sharing a similar multiset of features, differ in their root-dependent arrangement. Since node representations are computed based on rooted subtrees, treating such subtrees as identical ignores the positional context of features relative to the root. This oversight weakens the model’s ability to distinguish different nodes, thereby reducing the expressiveness of both node-level and graph-level representations. Consequently, the model becomes more prone to false positives during matching.

\subsection{Lack of Discriminative Power in The Scoring Stage}
To detect violations of containment constraints, prior work adopts the hinge distance measure $\Psi$, which can be formulated as follows:
\begin{align}
       \begin{cases}
          \Psi(Q,D) = \sum_d[(\mathbf{M}_Q - \mathbf{M}_D)_d]_+\\
          \Psi(v,u) = \sum_d[(\mathbf{M}_v - \mathbf{M}_u)_d]_+\\
       \end{cases}
\end{align}
To remain consistent with the well-established containment property formalized in Observation 1 Eq. (\ref{eq:1}), where the difference between any $D$ and $Q$ is element-wise non-negative, and the sum of the entries is also non-negative, $[\bullet]_+= \operatorname{Max}(0,\bullet)$ is used to enforce non-negative values. These measures operate within the $[0,+\infty)$ range, reflecting the graph-level and node-level subgraph edit distances, respectively. $(\mathbf{M}_Q - \mathbf{M}_D)_i > 0$ indicates that $Q$ is not contained by $D$ in the embedding space at dimensionality $i \in d$, thus violating the containment constraint. If $Q\subseteq D$, there should be no violations, resulting in $\Psi(Q,D) = 0$. We illustrate the hinge distance measure in Fig. \ref{fig:key} (Right).

Although they can evaluate compliance with the containment constraint, these measures have three main shortcomings. Firstly, they lack a clear reference point, making it difficult to consistently compare the distance scores. Secondly, a single extreme value can potentially distort the distances between other graphs. Furthermore, assigning all matched pairs a zero distance fails to distinguish their relative similarities and precludes applications that require ranking.
\section{The Proposed Architecture}
To alleviate the influence of scale differences in the encoding stage, our proposed NC-Iso preserves the relative positions between features within subtrees. Since such preservation can be reduced to maintaining the hierarchical dependencies between adjacent echelons of subtrees, the $k$-hop relative position consistency can be accordingly reduced to hierarchy consistency. NC-Iso treats the representation of root nodes from the previous layer and the aggregated information from neighborhoods as sequences, building hierarchical dependencies within subtrees and ensuring that matched graph pairs maintain consistent hierarchies. To solve the problem of existing measures, NC-Iso introduces a novel similarity measure. This measure normalizes the hinge distance as a compliance score to alleviate the influence of extreme values, then utilizes the representations of data graphs as reference points and quantifies the similarity and dissimilarity between graph pairs, thus effectively enabling the model the ability to rank matched pairs by considering the similarity dominance ratio (SDR).

\subsection{Hierarchy-aware GNN Encoder for Containment Constraint}

\textbf{Neighborhood Combination}. We first adopt a Linear layer to convert one-hot node labels into continuous space. 
To enable the hierarchy awareness of our GNN encoder, we propose modeling the hierarchical dependencies within node-rooted subtrees at the combination step during the message-passing process. An intuitive illustration can be found in Fig. \ref{fig:key} (Left). The $j$-th layer of the proposed architecture updates node representations as follows. 
\begin{align}
    \mathbf{M}^j_v = \phi^j(\operatorname{GRU}^{j}(\operatorname{Aggr^j}(\mathbf{M}_{S(v^\prime)^{j-1}}:v^\prime \in \mathcal{N}(v)), \mathbf{M}^{j-1}_v))\notag
\end{align}
Our proposed architecture utilizes a Gated Recurrent Unit (GRU) as the combination function. The GRU takes aggregated information, the embedding of $v$'s subtrees $S(v^\prime)$, as the input, and the representation of the root node from the previous layer as the hidden state. By considering each echelon of a node-rooted subtree as sequential elements, NC-Iso builds the hierarchical dependencies within the subtrees in a top-to-bottom manner. Furthermore, a two-layer MLP $\phi^j$ serves as a learnable hash function to ensure injectiveness of feature projection, following the approach suggested in \cite{Xu2019HowPA}. 

\textbf{Graph Summarization}.\label{sec:summarization} Since query graphs are smaller or have fewer nodes/edges than data graphs, 
we propose summarizing graphs with the Max operator, as it focuses on capturing the most salient features and is indifferent to graph scale, whereas the Sum and Mean operators can suffer from scale differences between graph pairs. 
Given the multi-scale node representation generated by the GNN backbone, the graph summarization can be expressed as follows.
\begin{align}
    \mathbf{M}^j_{G} = \operatorname{Max}(\mathbf{M}_v^j:v\in \mathcal{V}_G)\notag\\
    \mathbf{M}_{G} = \operatorname{Mean}(\mathbf{M}^j_{G}:j\leq k) \notag
\end{align}
Where $k$ represents the number of layers. We denote this entire encoding stage as $\Phi(\cdot)$.

\subsection{Similarity Dominance Ratio (SDR) Enhanced Measure for Scoring} Given graph representations of input pairs, the next stage is to detect violations of containment in the embedding space. However, previous measures, ranging from 0 to infinity, struggle to handle extreme values, provide an inconsistent comparison of distance scores, and lack discriminative power for matched pairs. To address these drawbacks, we put forth a novel distance measure, as illustrated in Fig. \ref{fig:key} (Right). This measure first transforms the hinge distance, signifying the violations of containment, into a compliance score as follows.
\begin{equation}
    Compliance(Q, D) = \exp(-\sum_d||[(\Phi(\mathbf{X}_{Q})-\Phi(\mathbf{X}_{D}))_d]_+||_2)\notag
\end{equation}
Where $exp(-x)$ normalize hinge distance by projecting $x\in [0,\infty)$ to the interval $(0,1]$. This normalized form alleviates the influence of extreme values and can be interpreted as a measure of the relative strength of compliance. 
To improve the ranking ability of NC-Iso for matched pairs, we first define two auxiliary functions: 
\begin{align}
&Inter(Q,D) = Min(\Phi(\mathbf{X}_{Q}),\Phi(\mathbf{X}_{D}))_d\notag\\
&Convex(Q,D) = Max(\Phi(\mathbf{X}_{Q}),\Phi(\mathbf{X}_{D}))_d\notag
\notag
\end{align}
The function $Inter(Q,D)$ estimates the intersection between multisets of graph pairs, while $Convex(Q,D)$ calculates the smallest multiset that contains both multisets of $Q$ and $D$ in the embedding space. Based on these functions, we compute the SDR as follows.
\begin{equation}
SDR(Q,D) = \frac{Inter(Q,D)}{\Phi(\mathbf{X}_{D})} - \frac{Convex(Q,D)-\Phi(\mathbf{X}_{D})}{Convex(Q,D)}\notag\\ 
\end{equation}
The $SDR$ incorporates the representation of the data graph as a reference point. 
The first term of $SDR$ assesses the proportion of common elements between the query $Q$ and the reference graph $D$, measuring their similarity. The second term focuses on the normalized dissimilarity of $Q$ with respect to $D$. It quantifies the difference between the multisets $\mathcal{M}_Q$ and $\mathcal{M}_D$, while ensuring that unmatchable pairs are not assigned a zero score in cases where $|\mathcal{M}_Q\cap\mathcal{M}_D| = 0$. By subtracting dissimilarity from similarity, we quantify the dominance ratio of similarity. 
The range of $SDR(Q,D)$ is $[-1, 1]$, representing $Q$ being completely different from $D$, or $Q$ being identical to $D$, respectively. Our proposed similarity measure can summarized as follows.
\begin{equation}
\Psi(Q,D) = Compliance(Q, D) \cdot SDR(Q,D)
\end{equation}
$\Psi(Q,D)$ overcomes the shortage of the previous hinge distance measures. It normalizes the violation score to the compliance score with $Compliance(\cdot, \cdot)$ to handle extreme values. It uses $SDR(\cdot,\cdot)$ to build reference points and depict the dominance ratio of similarity, which empowers the learned model with the ability to rank matched pairs. Once the model is trained, $Compliance(\cdot, \cdot)$ and $SDR(\cdot,\cdot)$ within the measure can be used separately to adapt to specific downstream tasks, providing a more flexible evaluation of subgraph isomorphism.

\textbf{Training and Predicting}. We train our model with the Mean Squared Error (MSE) Loss. It minimizes the distance between the predicted scores and the given scores $l$ of graph pairs, facilitating the control of the score range of our proposed measure.
\begin{align}
    \mathcal{L} = \operatorname{MSE}(\Psi(Q,D),l_{Q,D})
\end{align}
Given the predicted score of each pair, we determine subgraph matching via a threshold $\tau$ that is learned by a prediction function $p(\cdot)$ following \cite{Ying2020NeuralSM}, we adopt this function for every neural baseline to predict consistently:
\begin{align}
    p(Q,D)=
    \begin{cases}
    1\quad\text{iff}\quad \Psi(Q,D) > \tau,\\  
    0\quad otherwise.
    \end{cases}
\end{align}

\begin{algorithm}[hb]
\footnotesize
    \caption{ The NC-Iso algorithm}\label{alg:algo}
\begin{algorithmic}[1]
\REQUIRE {Data graph $D=(\mathbf{X}_D,\mathcal{A}_D)$, query graph $Q=(\mathbf{X}_Q,\mathcal{A}_Q)$, a preprocessor $\operatorname{Linear}_{pre}(\cdot)$, a $k$-layer $\operatorname{GNN}(\cdot)$ encoder, an $2$-layer $\operatorname{MLP}(\cdot)$, an $\operatorname{Linear}(\cdot)$ layer, a scoring function $\Psi(\cdot)$ and a predictor $p(\cdot)$}
\ENSURE{Is $Q$ isomorphic to a subgraph of $D$}
\STATE $\mathbf{H}^0_D = \operatorname{Linear}_{pre}(\mathbf{X}_D)\in\mathbb{R}^{|\mathcal{V}|_D\times 2d}$
\STATE $\mathbf{H}^0_Q = \operatorname{Linear}_{pre}(\mathbf{X}_Q)\in\mathbb{R}^{|\mathcal{V}|_D\times 2d}$
\FOR{$l = 1,\dots,K$}
\STATE $\mathbf{H}_v^l = \operatorname{GNN}(\mathbf{H}_v^{l-1}:v\in \mathcal{V}_D,\mathcal{A}_D)\in\mathbb{R}^{|\mathcal{V}_D|\times 2d}$
\STATE $\mathbf{H}_u^l = \operatorname{GNN}(\mathbf{H}_u^{l-1}:u\in \mathcal{V}_Q,\mathcal{A}_Q)\in\mathbb{R}^{|\mathcal{V}_D|\times 2d}$
\ENDFOR
\STATE $\mathbf{H}_v = \operatorname{MLP}(\{\mathbf{H}_v^{1},\dots,\mathbf{H}_v^K\})\in\mathbb{R}^{|\mathcal{V}_D|\times K \times d}$
\STATE $\mathbf{H}_u = \operatorname{MLP}(\{\mathbf{H}_u^{1},\dots,\mathbf{H}_u^K\})\in\mathbb{R}^{|\mathcal{V}_D|\times K \times d}$
\STATE $\mathbf{H}_D = \operatorname{Linear}(\operatorname{Max}(\mathbf{H}_v:v\in\mathcal{V}_D))\in\mathbb{R}^{K\times d}$
\STATE $\mathbf{H}_Q = \operatorname{Linear}(\operatorname{Max}(\mathbf{H}_u:u\in\mathcal{V}_Q))\in\mathbb{R}^{K\times d}$
\STATE $\bar{\mathbf{H}}_D=\operatorname{Mean}(\mathbf{H}^l_D:l<K)\in\mathbb{R}^d$
\STATE $\bar{\mathbf{H}}_Q=\operatorname{Mean}(\mathbf{H}^l_Q:l<K)\in\mathbb{R}^d$ \# End of Encoding Stage $\Phi(\cdot)$

\STATE score = $\Psi(\bar{\mathbf{H}}_D,\bar{\mathbf{H}}_Q) \in [0.0,1.0] $ \# End of Scoring Stage $\Psi(\cdot,\cdot)$ 
\STATE return $\operatorname{p}(score)$ \# Returning Prediction Based on Score 
\end{algorithmic}
\end{algorithm}

\begin{figure}[t]
\centering
\includegraphics[width=\linewidth]{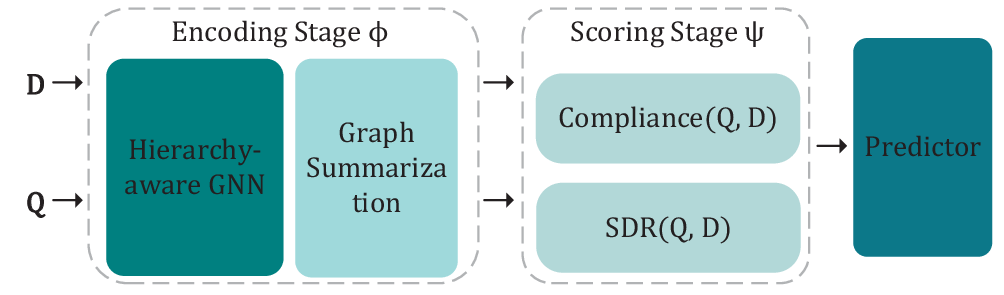}
\caption{The overview of NC-Iso.} \label{fig:pipeline}
\vspace{-0.5cm}
\end{figure}

\subsection{The Overview and Complexity Analysis}
\textbf{Overview.} The overview and pseudo-code of our proposed model can be found in Figure \ref{fig:pipeline} and Algorithm \ref{alg:algo}, respectively.

\textbf{Complexity Analysis.}
The complexity of generating graph representation via GNN is $\mathcal{O}(K|\mathcal{E}|d)$, where $K$ is the number of layers, $d$ is the dimensionality of representation, and $|\mathcal{E}|$ is the edge count in the graphs. Due to the usage of GRU, the complexity of our proposed architecture at each combination step is $\mathcal{O}(2d^2)$, thus the total complexity for the encoding stage is $\mathcal{O}(K(|\mathcal{E}|d+2d^2))$, and for the graph summarization, it is $\mathcal{O}(|\mathcal{V}|d)$. NC-Iso is efficient in subgraph matching prediction, whose per-prediction complexity is $\mathcal{O}(d)$.
\section{Evaluations} \label{sec:evaluation}
We compare NC-Iso against eight neural baselines and seven exact subgraph matching algorithms, considering: 
\begin{itemize}
    \item \emph{Effectiveness}: We assess the effectiveness of our model against neural baselines as our main result.
    \item \emph{Generalization ability}: We train models on small graphs and analyze their generalization ability to larger graphs.
    \item \emph{Scalability}: We train and evaluate models' performance and the impact of query sizes on models over large graph with one million nodes.
    \item \emph{Efficiency}: We compare NC-Iso's runtime with baselines.
    \item \emph{Ablation Study}: We investigate the impact and transferability of our proposed architecture and similarity measure.
    \item \emph{Hyperparameter Sensitivity}: We analyze the hyperparameter sensitivity of our proposed model in terms of number of layers and number of dimensionality.
    \item \emph{Case Study}: We visually analyze pairwise comparisons to gain a deeper understanding of the model's performance and behavior.
\end{itemize}

\begin{table*}[h]
\caption{The AUROC and Accuracy for the subgraph matching task over five runs with standard deviations. We mark the \textbf{best} and the \underline{second} performers. 'OOM' denotes out of memory.}\label{tab:main}
\resizebox{\linewidth}{!}{
\begin{tabular}{lcccccccccc}
\toprule
Dataset &
\multicolumn{2}{c}{AIDS} &
\multicolumn{2}{c}{COX2} &
\multicolumn{2}{c}{ENZYMES} &
\multicolumn{2}{c}{PROTEINS} &
\multicolumn{2}{c}{MSRC\_21} \\
\midrule
\# graphs &
\multicolumn{2}{c}{2,000} &
\multicolumn{2}{c}{467} &
\multicolumn{2}{c}{600} &
\multicolumn{2}{c}{1,113} &
\multicolumn{2}{c}{563} \\
\# node labels &
\multicolumn{2}{c}{38} &
\multicolumn{2}{c}{35} &
\multicolumn{2}{c}{3} &
\multicolumn{2}{c}{3} &
\multicolumn{2}{c}{24} \\
Avg. \# nodes &
\multicolumn{2}{c}{15.69} &
\multicolumn{2}{c}{41.22} &
\multicolumn{2}{c}{32.63} &
\multicolumn{2}{c}{39.06} &
\multicolumn{2}{c}{77.52} \\
Avg. \# edges &
\multicolumn{2}{c}{16.2} &
\multicolumn{2}{c}{43.45} &
\multicolumn{2}{c}{62.14} &
\multicolumn{2}{c}{72.82} &
\multicolumn{2}{c}{198.32} \\
\midrule
Baselines & AUROC $\uparrow$
& Acc $\uparrow$
& AUROC $\uparrow$
& Acc $\uparrow$
& AUROC $\uparrow$
& Acc $\uparrow$
& AUROC $\uparrow$
& Acc $\uparrow$
& AUROC $\uparrow$
& Acc $\uparrow$ 
\\ \midrule
GMN-embed \cite{Li2019GraphMN} &
$82.55 \pm 13.90$ &
$71.19 \pm 19.70$ &
$84.2 \pm 3.69$ &
$76.23 \pm 4.02$ &
$68.74 \pm 14.42$ &
$60.46 \pm15.25$ &  
$76.45 \pm9.53$ &
$67.35 \pm 12.12$ &
$87.94 \pm 12.69$ &
$71.22 \pm 29.09$ \\
SimGNN \cite{Bai2019SimGNNAN} &
$95.76 \pm 0.15$ &
$89.49 \pm 0.28$ &
$92.79 \pm 0.18$ &
$85.44 \pm 0.11$ &
$91.54 \pm 0.31$ &
$83.48 \pm 0.37$ &  
$93.27 \pm 0.77$ &
$85.68 \pm 1.05$ & 
\underline{$99.08 \pm 0.08$} &
\underline{$96.99 \pm 0.19$} \\
NeuroMatch \cite{Ying2020NeuralSM} &
$91.13 \pm 0.37$ &
$88.43 \pm 0.55$ &
$69.44 \pm 14.06$ &
$66.37 \pm 12.97$ &
$90.06 \pm 0.70$ &
$84.94 \pm 0.88$ & 
$92.10 \pm 0.61$ &
$86.82 \pm 0.89$ &  
$97.86 \pm 0.12$ &
$94.79 \pm 1.43$ \\
IsoNet \cite{Roy2022InterpretableNS} &
$53.89 \pm 0.63$ &
$51.09 \pm 1.63$ &
$78.96 \pm 0.81$ &
$50.10 \pm 0.13$ &
$57.43 \pm 0.68$ &
$50.71 \pm 1.05$ &   
OOM &
OOM &
$52.93 \pm 1.24$ &
$50.00 \pm 0.00$ \\
MCSNet \cite{roy2022maximum} &
$94.92 \pm 0.99$ &
$88.22 \pm 1.58$ &
$87.45 \pm 4.14$ &
$79.38 \pm 4.52$ &
$93.89 \pm 1.37$ &
$85.98 \pm 1.81$ & 
OOM &
OOM &
$98.80 \pm 0.07$ &
$96.99 \pm 0.37$ \\
Greed \cite{ranjan2022greed} &
$86.50 \pm 0.91$ &
$84.12 \pm 1.39$ &
$83.92 \pm 2.30$ &
$77.93 \pm 2.95$ &
$78.76 \pm 10.11$ &
$71.78 \pm 8.72$ &  
$68.29 \pm 12.99$ &
$64.89 \pm 9.65$ & 
$89.26 \pm 13.23$ &
$86.52 \pm 12.07$ \\
Eric \cite{eric} &
\underline{$97.39 \pm 0.11$} &
\underline{$92.32 \pm 0.25$} &
\underline{$93.40 \pm 0.75$} &
\underline{$86.16 \pm 0.86$} &
\underline{$94.67 \pm 0.40$} &
\underline{$87.37 \pm 0.68$} & 
\underline{$95.56 \pm 0.74$} &
\underline{$88.89 \pm 1.06$} & 
$98.76 \pm 0.18$ &
$96.31 \pm 0.36$ \\
D2Match \cite{liu2023d2match} &
$91.66 \pm 0.97$ &
$84.28 \pm 1.33$ &
$87.54 \pm 1.21$ &
$79.71 \pm 1.00$ &
$91.29 \pm 0.40$ &
$83.47 \pm 0.47$ &
$92.89 \pm 0.81$ &
$85.71 \pm 1.12$ &
$97.72 \pm 0.53$ &
$94.37 \pm 0.91$ \\
\midrule
Ours &
\textbf{97.55 $\pm$ 0.31} &
\textbf{93.37 $\pm$ 0.51} &
\textbf{94.50 $\pm$ 0.39} &
\textbf{87.84 $\pm$ 0.47} &
\textbf{96.44 $\pm$ 0.25} &
\textbf{90.52 $\pm$ 0.43} &
\textbf{97.08 $\pm$ 0.27} &
\textbf{91.52 $\pm$ 0.53} &
\textbf{99.36 $\pm$ 0.06} &
\textbf{98.15 $\pm$ 0.06} \\
\bottomrule
\end{tabular}}
\vspace{-0.5cm}
\end{table*}

\subsection{Experimental Setup} 
All models are trained on a single RTX 3090 GPU in a server with an Intel Xeon Silver 4210 CPU.

\textbf{Dataset.} We conduct experiments over seven real-world datasets from different domains \cite{Morris+2020}, including chemistry (AIDS, COX2), biology (Enzymes, Proteins, DD), image processing (MSRC\_21), and point cloud (FIRSTMM\_DB). Details on the dataset can be found in Table \ref{tab:main}.

\textbf{Evaluation Metrics.} We utilize the \emph{Area Under the Receiver Operating Characteristic (AUROC)} to assess the models' ability to discriminate between matched and unmatched graph pairs. We also employ \emph{Accuracy} as a measure of overall correctness for the models. Additionally, to evaluate the ranking ability, we use \emph{Spearman's Rank Correlation Coefficient ($\rho$)} and \emph{Hit@K}. 

\textbf{Baselines.} We compare our proposed model with NSM approaches NeuroMatch \cite{Ying2020NeuralSM}, IsoNet \cite{Roy2022InterpretableNS}, and the most recent D2Match \cite{liu2023d2match}. We also include the most popular and most recent graph similarity computation (GSC) models, such as GMN-embed \cite{Li2019GraphMN}, SimGNN \cite{Bai2019SimGNNAN}, MCSNet \cite{roy2022maximum}, Greed \cite{ranjan2022greed} and Eric \cite{eric}. We use the official implementation and hyperparameters provided by the authors to ensure fair comparisons. 
Given a graph pair as the input, NeuroMatch predicts a violation score of subgraph matching. Similarly, graph similarity computation (GSC) baselines predict a score (GED, SED, or MCS values) for the target problem. Thus, we made changes following NeuroMatch. We added a Linear layer, referred to as the predictor in our paper, for each baseline and trained them using cross-entropy loss. This ensures that all baselines predict subgraph matching in a consistent manner. 
Following \cite{liu2023d2match}, we select seven conventional subgraph matching algorithms, including QuickSI \cite{QuickSI}, GraphQL \cite{GraphQL}, CFL \cite{CFL}, VF3 \cite{VF3}, DP-iso \cite{DP-iso}, CECI \cite{CECI}, LFTJ \cite{In-Memory} in efficiency evaluation.

\textbf{Experimental Protocol.} Following the prior work \cite{Ying2020NeuralSM}, we partitioned the raw graphs in datasets into training and test graphs at a ratio of 4:1, with 20\% of the training graphs serving as the validation graphs. The ground truths are computed either by VF3 \cite{VF3} or by RapidMatch \cite{rapid}. Each batch includes 64 triplets of $(D, Q^+, Q^-)$, and each epoch comprises 100 iterations. Each model is trained for ten epochs as a warm-up phase and then tested on the validation set every epoch. We use early stopping with a patience of 50 to prevent overfitting. 
For conventional algorithms, we set a timeout of 100 seconds for each pair and record the total runtime to find the first solution for each graph pair as their inference time.

\begin{table}[tb]
\caption{The stats of sampled test set used in effectiveness and transferability evaluations.}\label{tab:stats}
\centering
\resizebox{\linewidth}{!}{
\begin{tabular}{lcccccccc}
\toprule
& 
AIDS &
COX2 &
ENZYMES &
PROTEINS &
MSRC\_21 &
DD &
FirstMM\_DB \\
\midrule
Avg. \# nodes ($D$) &
21.98 & 
30.37 &   
28.78 &        
48.95 &     
53.98 &  
277.80 &  
1210.48 \\
Avg. \# nodes ($Q$) &
8.90 &   
11.42 &     
10.80 &        
16.21 &       
18.50 &
76.24 &    
351.66 \\
Avg. \# edges ($D$) & 
22.94 &   
31.76 &    
50.00 &  
84.84 &   
129.83 &  
682.19 &   
2592.63 \\
Avg. \# edges ($Q$) & 
8.37 &   
11.00 &   
16.15 &       
25.17 &      
36.21 &   
174.79&   
724.21 \\
\bottomrule
\end{tabular}}
\vspace{-0.5cm}
\end{table}

\textbf{Dataset Sampling Strategy.}
Based on validation and test graphs, we generated offline validation and test sets to ensure a fair comparison, while training sets were generated in an on-the-fly manner based on training graphs. We adopt the random walk-based sampling technique commonly used in NSM research, \cite{Ying2020NeuralSM,Roy2022InterpretableNS,liu2023d2match}, which randomly chooses a start node and then walks to one of the neighbors of the visited nodes till a length $n$, then extracts the subgraph induced by the visited nodes in the corresponding raw graph to generate data graphs. The matchable query is generated similarly based on the sampled data graph, while the unmatchable query is extracted from another graph. The validation set includes 1,024 batches, while the test set has 2,048 batches for each dataset, resulting in abundant amounts of graph pairs of 131,072 and 262,144, respectively. The statistics of the sampled test sets are available in Table \ref{tab:stats}.

\textbf{Implementation Details}
We maintained consistent hyperparameters across all datasets. Initially, we applied a Linear layer to convert the one-hot node labels into node features with a dimensionality of 64. Next, we employed a 6-layer GNN with GRU as the combination function. Each GNN layer had 64-dimensional inputs and outputs. To reduce the dimensionality of the node features to 32, we utilized a two-layer MLP with input dimensions [64,32] and output dimensions [32,32]. For inter-layer operations, we applied the ReLU activation function and layer normalization. Graph-level embeddings were generated by employing max pooling for each layer. Subsequently, a Linear layer was utilized to map the embeddings to the same feature space, with input and output dimensionalities of 32. Ultimately, we took the average of each dimension to obtain the final graph-level embedding. As for optimization, we employed Adam with a fixed learning rate of $10^-3$. Regarding the loss function, we first clamped the embeddings to ensure that each dimension is greater than or equal to a small positive value of $1e-7$. We set a maximum duration of 8 hours for each run due to the equipment limitation, excluding the graph generation time.

\subsection{Effectiveness}
We compare the AUROC and Accuracy between neural-based models while omitting conventional algorithms because they provide exact matches. 
The results of NC-Iso and the neural baselines are detailed in Table \ref{tab:main}. NC-Iso consistently showcases competitive performance across all datasets and evaluation metrics. Benefiting from the hierarchical dependency preservation and SDR-enhanced measure, NC-Iso achieves substantial improvements compared with NSM methods with an absolute increase of up to 6.95\% in AUROC and exhibits competitive performance against the most advanced GSC models using more sophisticated architectures. Although fine-grained node alignment methods like MCSNet generally perform better than coarse-grained graph-level approaches such as NeuroMatch, they are less scalable. Moreover, fine-grained alignment cannot guarantee better performance, as both the top 2 NC-Iso and Eric are coarse-grained approaches, while IsoNet, aligning edges to estimate cost, did not perform as well as expected. It may be due to the Gumbel-Sinkhorn network used in IsoNet forcing a one-to-one correspondence, which may not effectively handle partial matches of unmatchable pairs, generating similar scores for both matched and unmatched pairs.

\begin{table}[h]
\caption{The results for generalization ability evaluation.}\label{tab:trans}
\resizebox{\linewidth}{!}{
\begin{tabular}{lcccc}
\toprule
Dataset &
\multicolumn{2}{c}{DD} &
\multicolumn{2}{c}{FirstMM\_DB} \\
\midrule
Baselines & AUROC $\uparrow$
& Acc $\uparrow$
& AUROC $\uparrow$
& Acc $\uparrow$
\\ \midrule 
GMN-embed &
$65.16 \pm 6.06$ &
$52.08 \pm 9.97$ &
$55.81 \pm 1.05$ &
$50.75 \pm 1.36$ \\
SimGNN &
\underline{$93.27 \pm 1.90$} &
\underline{$86.14 \pm 2.04$} &
\underline{$83.56 \pm 4.33$} &
\underline{$76.94 \pm 5.57$} \\
NeuroMatch &
$77.27 \pm 1.38$ &
$69.93 \pm 2.94$ &
$80.22 \pm 3.71$ &
$78.56 \pm 3.50$ \\
Greed &
$59.78 \pm 8.10$ &
$58.11 \pm 5.22$ &
$63.64 \pm 6.29 $ &
$63.41 \pm 6.03$ \\
Eric &
$91.54 \pm 2.33$ &
$84.46 \pm 2.02$ &
$81.18 \pm 4.85$ &
$74.64 \pm 3.26$ \\
\midrule
Ours &
\textbf{98.81 $\pm$ 0.93} &
\textbf{94.59 $\pm$ 0.27} &
\textbf{91.71 $\pm$ 3.20} &
\textbf{86.33 $\pm$ 2.92} \\
\bottomrule
\end{tabular}}
\vspace{-0.5cm}
\end{table}

\subsection{Generalizing to Larger, Unseen Graphs}
Due to the NP-complete nature of subgraph matching, acquiring training data for larger graphs is challenging. Thus, it is crucial for neural-based models to be trained on small graphs and generalize to larger ones. In this experiment, all models were trained on graphs containing fewer than 100 nodes and subsequently evaluated on unseen graphs with up to five thousand nodes. The results do not include models that encountered out-of-memory issues. As presented in Table \ref{tab:trans}, our proposed NC-Iso exhibits an absolute improvement of up to 8\% in AUROC. While other coarse-grained methods, such as Eric and NeuroMatch, exhibit a substantial performance drop, our proposed method maintains superior performance. This may be because NC-Iso considers the influence of scale differences. NC-Iso incorporates hierarchical dependencies, ensures that matched graph pairs maintain consistent hierarchies, and uses max pooling to alleviate the influence of noise introduced by scale differences between graph pairs. 

\begin{table}[h]
\caption{The AUROC for subgraph matching on the large-scale dataset.}\label{tab:scal}
\centering
\resizebox{\linewidth}{!}{
\begin{tabular}{lcccccccccc}
\toprule
Dataset &
SimGNN &
NeuroMatch &
Greed &
Eric &
D2Match &
Ours \\
\midrule
OGB-Arxiv  &
OOM &
$73.25$ &
$51.32$ &
$56.01$ &  
\underline{74.27} &
\textbf{74.87} \\
Youtube &
OOM &
$75.82$ &
\underline{77.05} &
$72.47$ & 
OOM &
\textbf{93.26} \\
\bottomrule
\end{tabular}}
\vspace{-0.5cm}
\end{table}

\subsection{Scaling to Large-scale Graphs} 
We conducted experiments on large-scale datasets such as \emph{OGB-Arxiv} \cite{ogb} and \emph{Youtube} \cite{In-Memory}. Specifically, \emph{OGB-Arxiv} dataset contains a node-labeled graph with 169,343 nodes, 2,987,624 edges, and 40 label types, whereas \emph{Youtube} dataset contains a node-labeled graph with 1,134,890 nodes, 1,166,243 edges, and 25 label types. We train the models on small sampled data and query graphs, then evaluate sampled queries of different sizes on the original graphs. Since the full-graph representation learning for large graphs imposes a large memory requirement for GNN training and leads to an inefficient gradient update, we followed the setup of Greed \cite{ranjan2022greed}. This involved extracting the $k$-hop neighborhood $D_v$ around each node $v \in D$, then computing the representation of the $k$-hop neighborhoods as the node embeddings of the corresponding nodes. If the query $Q$ could match any node in $D$, we considered it a match. By comparing the query graph embedding with each node embedding in $D$, neural-based methods predict the subgraph matching. We evaluated each dataset using 20K graph pairs against selected baselines that have shown strong performance in transferability experiments or are highly related to our approach. The AUROC for subgraph matching on the large-scale dataset is reported in Table \ref{tab:scal}, which showcased the strong generalization ability and scalability of our proposed NC-Iso compared with selected approaches. 

\begin{table*}[tb]
\vspace{-0.3cm}
\caption{Impact of Query Graph Size}\label{tab:impact}
\centering
\resizebox{0.7\linewidth}{!}{
\begin{tabular}{l|cccccc|c}
\toprule
\multicolumn{2}{l}{\textbf{Dataset}}& 
SimGNN &
NeuroMatch &
Greed &
Eric &
D2Match &
Ours\\
\midrule
\multirow{5}{*}{Youtube} &
$[5, 20)$ &
OOM &
50.35 &
51.37 &
\underline{74.64} &
56.01 &
\textbf{89.43} \\
& $[20, 50)$ &
OOM &
66.98 &
70.35 &       
\underline{88.90} &
54.20 &
\textbf{90.89} \\
& $[50, 100)$ &
OOM &
90.87 &
91.11 &       
\underline{93.96} &
57.49 &
\textbf{94.81} \\
& $[100, 200)$ &
OOM &
94.28 &
\underline{95.14} &       
\textbf{95.88} &
66.30 &
94.26 \\
& $[200,500)$ &
OOM &
\textbf{96.89} &
\underline{96.75} &       
76.47 &
OOM &
96.18 \\
\midrule

\multirow{6}{*}{FirstMM\_DB} &
$[5, 20)$ &
\textbf{86.42} &
77.17 &
71.73 &
\underline{84.91} &
81.91 &
84.72 \\
& $[20, 50)$ &
\underline{87.13} &
76.99 &
74.03 &
86.07 &
83.70 &
\textbf{89.04} \\
& $[50, 100)$ &
\underline{86.09} &
76.07 &
72.28 &
84.56 &
85.39 &
\textbf{89.91} \\
& $[100,200)$ &
84.15 &
75.67 &
71.03 &
82.12 &
\underline{86.60} &
\textbf{89.28} \\
& $[200,500)$ &
76.98 &
73.57 &
67.93 &
71.22 &
\underline{83.77} &
\textbf{84.30} \\
\bottomrule
\end{tabular}}
\vspace{-0.5cm}
\end{table*}

\subsection{Impact of Query Graph Size}
This experiment analyzes the effectiveness of models on queries of different sizes. we compare with baselines shown competitive performance in large-scale experiments or are highly related to our approach. Intuitively, increasing the size difference between the data and query graphs introduces more noise factors that can affect predictions. As a result, smaller queries pose a more challenging task. Results of this experiment are shown in Table \ref{tab:impact}. our proposed method exhibits a substantial improvement on small queries with less than 100 nodes compared with the baselines of the same kind. And our proposed model also reaches an overall competitive performance compared with all baselines.

\subsection{Efficiency Compared with Baselines} \label{app:efficiency}
The runtime analysis, as shown in Table \ref{tab:infer}. NC-Iso exhibits a slightly slower runtime in milliseconds compared with another coarse-grained model, Greed. This discrepancy can be attributed to the use of GRU as the combination function in NC-Iso, which incurs higher computational costs than GIN in Greed. While the runtime of D2Match is comparable to that of SimGNN, it suffers from the lengthy preprocessing time that transforms the mutual cyclic features present in the data and query graphs into a super-node representation. The cost of this preprocessing step grows with the scale of the graphs, leading to potential efficiency challenges for larger graph sizes. Although the training time was omitted in our results, we found it generally aligns with the inference time for the models. Except for Eric, who stands out as the fourth fastest model in terms of inference time but is slow in training. This is because once Eric is trained, its matching model can be detached from the inference pipeline, resulting in improved inference efficiency. All neural-based methods are consistently faster than the considered conventional algorithms across all datasets. However, it is crucial to recognize that neural-based methods and conventional algorithms are oriented towards different scenarios. While conventional algorithms excel in achieving precise bijective mappings with guaranteed correctness, they rely on heuristic strategies, which may restrict their generalization ability. Moreover, their time consumption grows exponentially as graph size grows, and it becomes especially challenging for them to filter out unmatchable graphs, as this task requires exploring all possible combinations of subgraphs to demonstrate that no valid match exists. Furthermore, these conventional algorithms preprocess graph pairs in an on-the-fly manner, preventing the reuse of graph information. In contrast, the neural-based method depends on learned features, providing quick predictions with scores and producing reusable graph representations. These characterizations of neural-based methods make them more suitable for retrieval-related applications.

\begin{table*}[tb]
\caption{Average inference time per batch (in seconds) compared
with the baselines.}\label{tab:infer}
\centering
\resizebox{0.6\linewidth}{!}{
\begin{tabular}{lccccc}
\toprule
        & AIDS & COX2 & ENZYMES & PROTEINS & MSRC\_21 \\
\midrule
QuickSI \cite{QuickSI} &
$1.8298$ &
$1.2471$ & 
$1.1697$ & 
$2.4307$ &        
$2.5577$ \\
GraphQL \cite{GraphQL} &
$2.2719$ &
$1.4099$ &
$0.9752$ &
$1.3762$ &
$2.4765$ \\
CFL \cite{CFL} &
$2.3062$ &
$1.1325$ & 
$1.2016$ &  
$1.4316$ &    
$3.0425$ \\
DP-iso \cite{DP-iso} &
$2.5315$ &
$1.6045$ &
$1.1839$ &
$1.5358$ &
$3.6096$ \\
CECI \cite{CECI} &
$6.7468$ &
$6.9154$ &
$4.6092$ &
$5.5786$ & 
$8.3250$ \\
LFTJ \cite{In-Memory} &
$2.2281$ &
$1.1552$ &
$1.1512$ &
$1.0621$ & 
$2.2975$ \\
VF3 \cite{VF3} &
$0.2257$ &
$0.2265$ &
$0.2265$ &
$0.3173$ &
$0.2766$ \\
\midrule
GMN-embed \cite{Li2019GraphMN} &
$0.0044$ &
$0.0045$ & 
$0.0043$ & 
$0.0047$ &        
$0.0056$ \\
SimGNN \cite{Bai2019SimGNNAN} &
$0.0161$ &
$0.0151$ & 
$0.0156$ & 
$0.0167$ &        
$0.0154$ \\
NeuroMatch \cite{Ying2020NeuralSM} &
$0.0028$ &
$0.0028$ & 
$0.0028$ & 
$0.0030$ &        
$0.0031$ \\
IsoNet \cite{Roy2022InterpretableNS} &
$0.0483$ &
$0.0338$ & 
$0.0809$ & 
OOM &        
$0.4425$ \\
MCSNet \cite{roy2022maximum} &
$0.0628$ &
$0.0528$ & 
$0.0737$ & 
OOM &        
$0.0798$ \\
Greed \cite{ranjan2022greed} &
$0.0023$ &
$0.0021$ & 
$0.0020$ & 
$0.0021$ &        
$0.0022$ \\
Eric \cite{eric} &
$0.0044$ &
$0.0038$ & 
$0.0041$ & 
$0.0041$ &        
$0.0042$ \\
D2Match (Preprocess) \cite{liu2023d2match} &
$0.3780$ &
$0.4979$ & 
$0.7845$ & 
$1.7264$ &        
$2.0564$ \\
D2Match \cite{liu2023d2match} &
$0.0268$ &
$0.0271$ & 
$0.0288$ & 
$0.0346$ &        
$0.0382$ \\
\midrule
Ours &
$0.0024$ &
$0.0024$ & 
$0.0024$ & 
$0.0025$ &        
$0.0026$ \\
\bottomrule
\end{tabular}}
\end{table*}

\subsection{Ablation Study}

\begin{table*}[tb]
\caption{The AUROC and Accuracy over five runs on the effectiveness of proposed architecture.}\label{tab:full_ab}
\resizebox{\linewidth}{!}{
\begin{tabular}{lcccccccccc}
\toprule
Dataset &
\multicolumn{2}{c}{AIDS} &
\multicolumn{2}{c}{COX2} &
\multicolumn{2}{c}{ENZYMES} &
\multicolumn{2}{c}{PROTEINS} &
\multicolumn{2}{c}{MSRC\_21} \\
{} & AUROC $\uparrow$
& Acc $\uparrow$
& AUROC $\uparrow$
& Acc $\uparrow$
& AUROC $\uparrow$
& Acc $\uparrow$
& AUROC $\uparrow$
& Acc $\uparrow$
& AUROC $\uparrow$
& Acc $\uparrow$ 
\\ \midrule
Ours+GIN &
97.00 $\pm$ 0.08 &
92.38 $\pm$ 0.39 &
94.44 $\pm$ 0.17 &
87.28 $\pm$ 0.42 &
93.95 $\pm$ 0.67 &
86.42 $\pm$ 0.67 &
95.12 $\pm$ 0.60 &
88.56 $\pm$ 0.86 &
99.15 $\pm$ 0.09 &
97.80 $\pm$ 0.16 \\
Ours+GCN &
96.21 $\pm$ 0.24 &
91.03 $\pm$ 0.35 &
93.21 $\pm$ 0.54 &
86.76 $\pm$ 0.38 &
92.10 $\pm$ 0.56 &
84.27 $\pm$ 0.61 &
93.22 $\pm$ 0.77 &
85.95 $\pm$ 1.10 &
99.03 $\pm$ 0.07 &
97.32 $\pm$ 0.21 \\
Ours+SAGE &
96.23 $\pm$ 0.37 &
90.74 $\pm$ 0.53 &
94.26 $\pm$ 0.37 &
87.20 $\pm$ 0.51 &
93.50 $\pm$ 0.46 &
85.89 $\pm$ 0.73  &
94.30 $\pm$ 0.52 &
87.43 $\pm$ 0.84 &
98.97 $\pm$ 0.10 &
97.24 $\pm$ 0.12 \\
Ours+GAT &
96.10 $\pm$ 0.18 &
90.37 $\pm$ 0.39 &
93.27 $\pm$ 0.41 &
86.60 $\pm$ 0.29 &
91.14 $\pm$ 0.21 &
82.93 $\pm$ 0.37 &
92.20 $\pm$ 1.70 &
84.36 $\pm$ 2.17 &
99.03 $\pm$ 0.07 &
97.01 $\pm$ 0.23 \\
\midrule
Ours+Mean pool &
96.36 $\pm$ 0.19 &
90.60 $\pm$ 0.36 &
93.07 $\pm$ 0.34 &
85.67 $\pm$ 0.50 &
93.06 $\pm$ 0.48 &
85.87 $\pm$ 0.73 &
94.48 $\pm$ 0.65 &
87.57 $\pm$ 0.87  &
98.91 $\pm$ 0.07 &
96.35 $\pm$ 0.27 \\
Ours+Sum pool &
94.89 $\pm$ 0.42 &
88.42 $\pm$ 0.63 &
58.45 $\pm$ 18.85 &
57.03 $\pm$ 15.72 &
90.84 $\pm$ 0.57 &
83.06 $\pm$ 0.93 &
91.79 $\pm$ 0.84 &
84.09 $\pm$ 1.03  &
50.00 $\pm$ 0.00 &
50.00 $\pm$ 0.00 \\
\midrule
Ours &
\textbf{97.55 $\pm$ 0.31} &
\textbf{93.37 $\pm$ 0.51} &
\textbf{94.50 $\pm$ 0.39} &
\textbf{87.84 $\pm$ 0.47} &
\textbf{96.44 $\pm$ 0.25} &
\textbf{90.52 $\pm$ 0.43} &
\textbf{97.08 $\pm$ 0.27} &
\textbf{91.52 $\pm$ 0.53} &
\textbf{99.36 $\pm$ 0.06} &
\textbf{98.15 $\pm$ 0.06} \\
\bottomrule
\end{tabular}}
\vspace{-0.5cm}
\end{table*}

\paragraph{Effectiveness of Proposed Architecture}
To demonstrate the effectiveness of our backbone, which preserves the hierarchical dependencies within node-rooted subtrees, we compare it against popular GNNs such as GIN, GCN, GraphSAGE, and GAT. Additionally, We replace the Max operator in Section \ref{sec:summarization} with the Sum and Mean operators to validate our intuition. We report the results in Table \ref{tab:full_ab}. We observe an improvement of up to $3\%$ in AUROC, highlighting the effectiveness of our proposed architecture. Results show that our proposed architecture consistently outperforms these GNN backbones that utilize permutation-invariant combination functions, as they do not inherently preserve the hierarchical dependencies within node-rooted subtrees. Moreover, the Max operator is an optimal choice compared with the Sum and Mean operations for our proposed architecture, which aligns with our design intuition. Since query graphs are smaller or have fewer nodes/edges than data graphs, additional node or edge representations in $D$ may introduce irrelevant information. The Sum operator treats both relevant and irrelevant features equally, potentially resulting in the accumulation of noise. This is particularly problematic for the MSRC\_21 dataset, where prediction performance under Sum pooling drops to near-random levels. We attribute this to the fact that MSRC\_21 contains a significantly higher number of triangular motifs, forming mesh-like graph structures. In subgraph matching tasks, such local connectivity may generate many structurally similar subregions. Under these conditions, Sum pooling tends to dilute discriminative features due to excessive aggregation, causing loss of important structural distinctions. The Mean operator, in turn, may amplify the influence of irrelevant elements. By prioritizing important features, the Max operator mitigates the impact of irrelevant information and leads to a more robust representation of the graphs.

\begin{figure}[tb]
\includegraphics[width=1.0\linewidth]{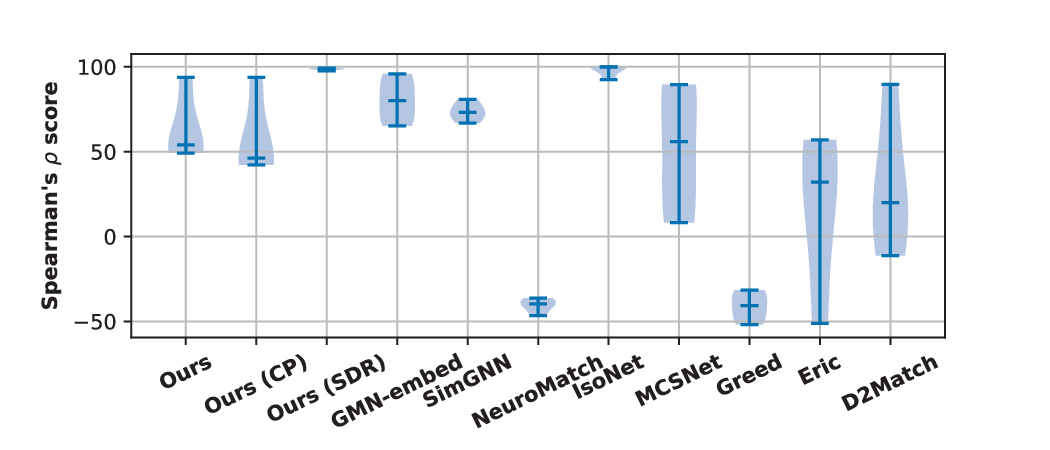}
\caption{Ranking results on matched pairs.}\label{fig:ranking}
\vspace{-0.5cm}
\end{figure}

\paragraph{Effectiveness and Flexibility of Proposed Measure} 
In this experiment, we evaluate the effectiveness and flexibility of individual components within our proposed measure, including $Compliance (CP)$, which assesses constraint compliance, $SDR$, considering the dominance of similarity, and their combination $\Psi$, on two downstream tasks: 

\textbf{Ranking Matched Pairs}: 
We evaluated the models' ability to \emph{rank matched graphs}. We sampled 20 matched queries per data graph, ensuring subgraph isomorphism transitivity. Median and range of $\rho$ scores across four datasets are reported due to 'OOM' errors of MCSNet and IsoNet on the Proteins dataset. The ranking results, depicted in Fig. \ref{fig:ranking}, reveal that our proposed $SDR$ reliably ranks matched pairs with the highest and the most stable $\rho$ score, aligning with our design, outperforming $CP$ that focuses solely on constraint compliance, and $\Psi$ that considers both compliance and similarity. This advantage makes SDR particularly valuable in retrieval-oriented tasks or top-$k$ candidate filtering, where relative ranking of subgraph candidates is critical. Notably, even though all three methods focus on constraint compliance, our $CP$ significantly improves the ranking of matched pairs compared with NeuroMatch and Greed. This improvement may result from using data graph representations as reference points in our measure, which allows $CP$ to provide more consistent score comparisons. IsoNet achieves the second-best ranking of matched pairs, benefiting from the iterative refinement of the Gumbel-Sinkhorn network while suffering from its expensive computational cost.

\begin{table*}[tb]
\caption{Effectiveness of node-level alignment, we report the average $Hit@k$ per node.}\label{tab:align}
\centering
\resizebox{0.7\linewidth}{!}{
\begin{tabular}{lcccccccccc}
\toprule
Dataset &
\multicolumn{2}{c}{AIDS} &
\multicolumn{2}{c}{COX2} &
\multicolumn{2}{c}{ENZYMES} &
\multicolumn{2}{c}{PROTEINS} &
\multicolumn{2}{c}{MSRC\_21} \\
{} &
$Hit@1$ &
$Hit@3$ &
$Hit@1$ &
$Hit@3$ &
$Hit@1$ &
$Hit@3$ &
$Hit@1$ &
$Hit@3$ &
$Hit@1$ &
$Hit@3$ \\
\midrule
Greed &
$50.73$ &
$81.98$ &
$47.03$ &
$71.04$ &  
$31.77$ &
$58.17$ &
$35.40$ &
$60.58$ &
$30.53$ &  
$49.26$ \\
NeuroMatch &
\underline{70.81} &
$89.37$ &
$47.88$ &
\underline{71.62} &  
$34.64$ &
$59.51$ &
$41.58$ &
$64.50$ &
$20.78$ &  
$37.49$ \\
\midrule
Ours (SDR) &
$65.25$ &
$78.53$ &
$53.20$ &
$68.24$ &  
$36.76$ &
$54.74$ &
$49.71$ &
$64.71$ &
$42.86$ &  
$60.05$ \\
Ours (CP) &
$70.22$ &
\underline{90.28} &
\underline{59.14} &
\textbf{77.38} &  
\underline{42.38} &
\underline{68.78} &
\underline{53.53} &
\underline{74.26} &
\underline{44.33} &  
\underline{68.28} \\
Ours &
\textbf{77.69} &
\textbf{92.07} &
\textbf{59.16} &
\textbf{77.38} &  
\textbf{43.01} &
\textbf{69.06} &
\textbf{54.95} &
\textbf{74.94} &
\textbf{47.13} &  
\textbf{69.22} \\
\bottomrule
\end{tabular}}
\vspace{-1.5mm}
\end{table*}

\textbf{Node-level Alignment}: In line with previous research, we considered subgraph matching a graph-level task and utilized graph-level embedding for prediction. However, our proposed measure and each of its components can also be applied at the node level to provide insights for \emph{node alignment}. To validate this ability, we compared each query-data node pair and reported the $Hit@K$ of 12,800 graph pairs. 

We compare our proposed model with NeuroMatch and Greed, as they are graph representation models similar to ours, and they predict subgraph matching with hinge distance measures. The results are shown in Table \ref{tab:align}, which clearly demonstrate the significant improvement of NC-Iso compared with the baselines of the same kind. 
Combining $CP$ and $SDR$ together can improve the performance in node-level alignment, suggesting that our proposed architecture and distance measure can also be applied to node-level NSM tasks with proper supervision, which we leave for future work. These two experiments show that our similarity measure can be flexibly utilized to address different downstream tasks related to subgraph matching.

\begin{figure}[!ht]
\centering
   \begin{minipage}{0.40\linewidth}{
   \includegraphics[width=1\linewidth]{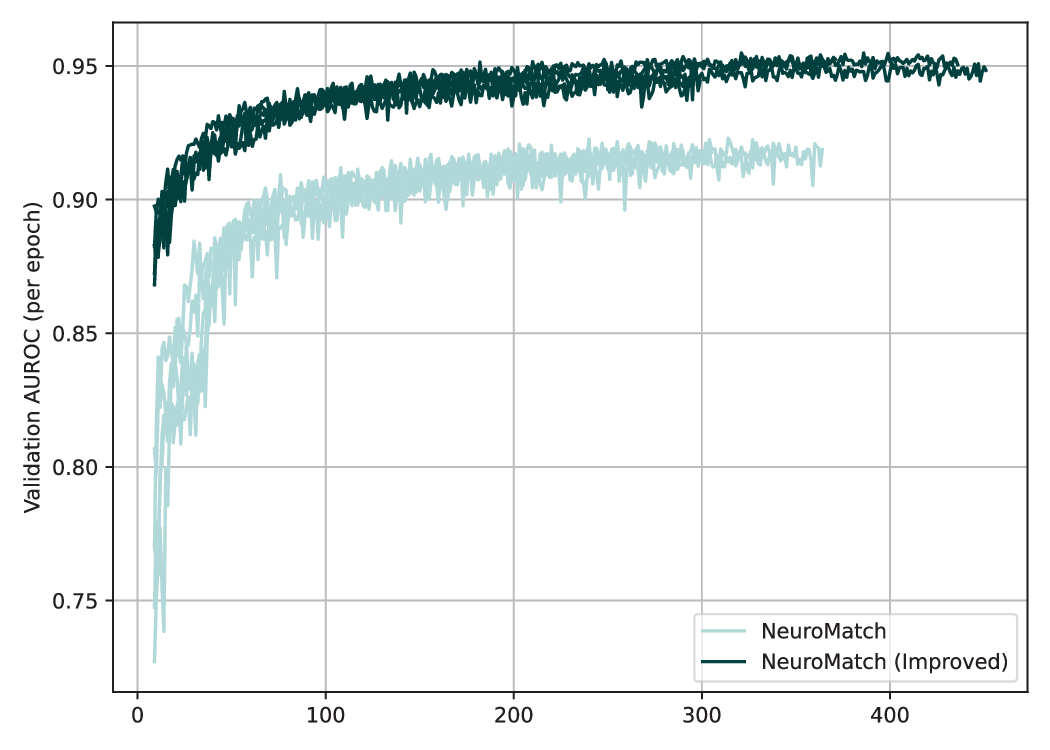}}
   \end{minipage}
   \begin{minipage}{0.40\linewidth}{
   \includegraphics[width=1\linewidth]{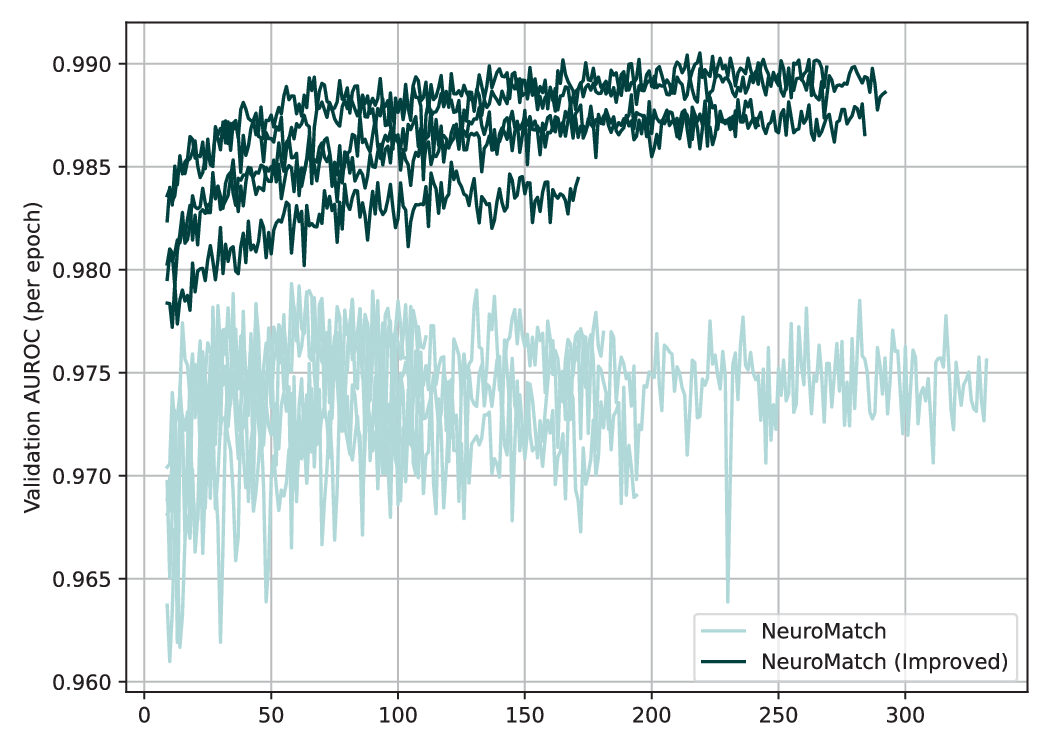}}
   \end{minipage}
   \caption{We alter the hinge distance measure in NeuroMatch with our proposed one. The validation AUROC on PROTEINS (Left) and MSRC\_21 (Right) datasets demonstrate the substantial improvement brought by our proposed measure.}\label{fig:plugin}
   \vspace{-0.5cm}
\end{figure}

\textbf{Transferability of Proposed Components.}
We conducted two experiments to validate the proposed components' transferability. Firstly, we replaced the backbone and jumping connection employed in NeuroMatch and Greed with our proposed GNN backbone. Secondly, we incorporated our measure into NeuroMatch, replacing its original smoothed hinge distance measure. The results, as shown in Table \ref{tab:trans_comp}, indicate that our proposed backbone improves the stability of their models, and our proposed measure leads to a substantial performance boost. The training dynamics of the enhanced NeuroMatch model, which incorporates the proposed similarity measure, and the original NeuroMatch model are depicted in Fig. \ref{fig:plugin}. The results provide evidence of the notable enhancement achieved by our proposed measure. The performance improvement is evident from the early stages of training for NeuroMatch, resulting in an improved model performance in terms of AUROC compared with the original model.
\begin{table*}[tb]
\caption{Transferability study of the proposed components. We present the AUROC.}\label{tab:trans_comp}
\centering
\resizebox{0.7\linewidth}{!}{
\begin{tabular}{lcccccccccc}
\toprule
Dataset &
AIDS &
COX2 &
ENZYMES &
PROTEINS &
MSRC\_21 \\
\midrule
Greed &
$86.50 \pm 0.91$ &
$83.92 \pm 2.30$ &
$78.76 \pm 10.11$ &
$68.29 \pm 12.99$ &  
$89.26 \pm 13.23$ \\
Greed+Backbone &
$86.52 \pm 1.81$ &
$83.26 \pm 1.28$ &
$84.68 \pm 0.81 $ &
$85.60 \pm 0.81$ &  
$95.36 \pm 0.37$ \\
\midrule
NeuroMatch &
$91.13 \pm 0.37 $ &
$69.44 \pm 14.06$ &
$90.06 \pm 0.70$ &  
$92.10 \pm 0.61$ &
\underline{$97.86 \pm 0.12$} \\
NeuroMatch+Backbone &
\underline{$91.17 \pm 0.72$} &
\underline{$84.94 \pm 1.82$} &
\underline{$90.46 \pm 0.45$} &
\underline{$92.17 \pm 0.44$} &  
\underline{$97.86 \pm 0.11$} \\
NeuroMatch+Enhanced measure &
\textbf{96.90 $\pm$ 0.11} &
\textbf{93.41 $\pm$ 0.18} &
\textbf{94.17 $\pm$ 0.30} &
\textbf{95.16 $\pm$ 0.52} &
\textbf{98.90 $\pm$ 0.10} \\
\bottomrule
\end{tabular}}
\vspace{-3.0mm}
\end{table*}

\begin{figure}[!ht]
   \begin{minipage}{0.49\linewidth}{
   \includegraphics[width=1\linewidth]{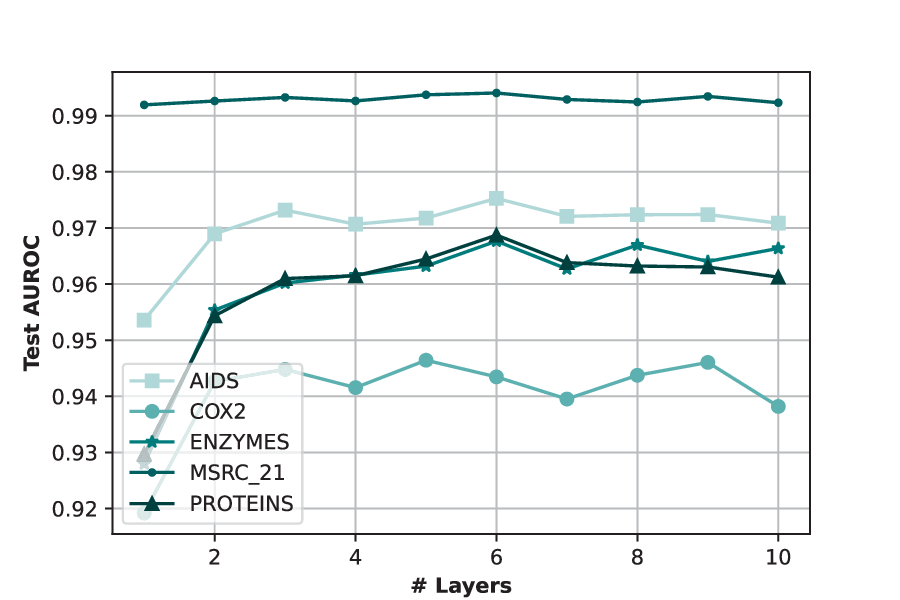}}
   \end{minipage}
   \begin{minipage}{0.49\linewidth}{
   \includegraphics[width=1\linewidth]{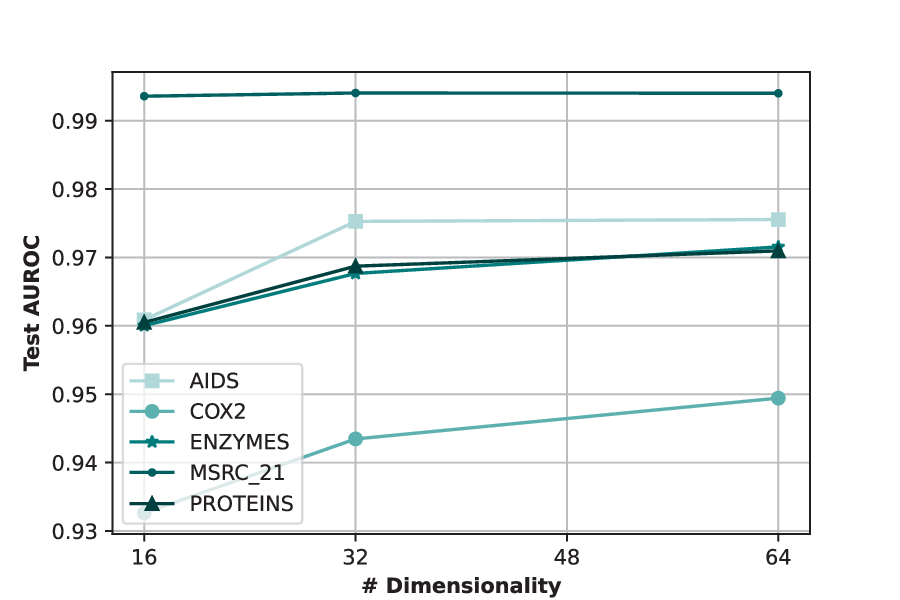}}
   \end{minipage}
   \caption{Hyperparameter sensitivity analysis of NC-iso. Sensitivity on \# Layers (Left). Sensitivity on \# Dimensionality (Right).}\label{fig:sens}
   \vspace{-0.5cm}
\end{figure}

\subsection{Hyperparameter Sensitivity} \label{app:sensitivity}
In order to analyze the hyperparameter sensitivity of NC-Iso, we conducted experiments on the number of layers and the number of dimensionality, as illustrated in Figure \ref{fig:sens}. 
Results show that NC-iso is more responsive to changes in dimensionality than the number of layers, the impact of latter is likely due to the dataset's specific characteristics. 
\begin{figure}[tb]
   \begin{subfigure}{0.49\linewidth}{
   \includegraphics[width=1\linewidth]{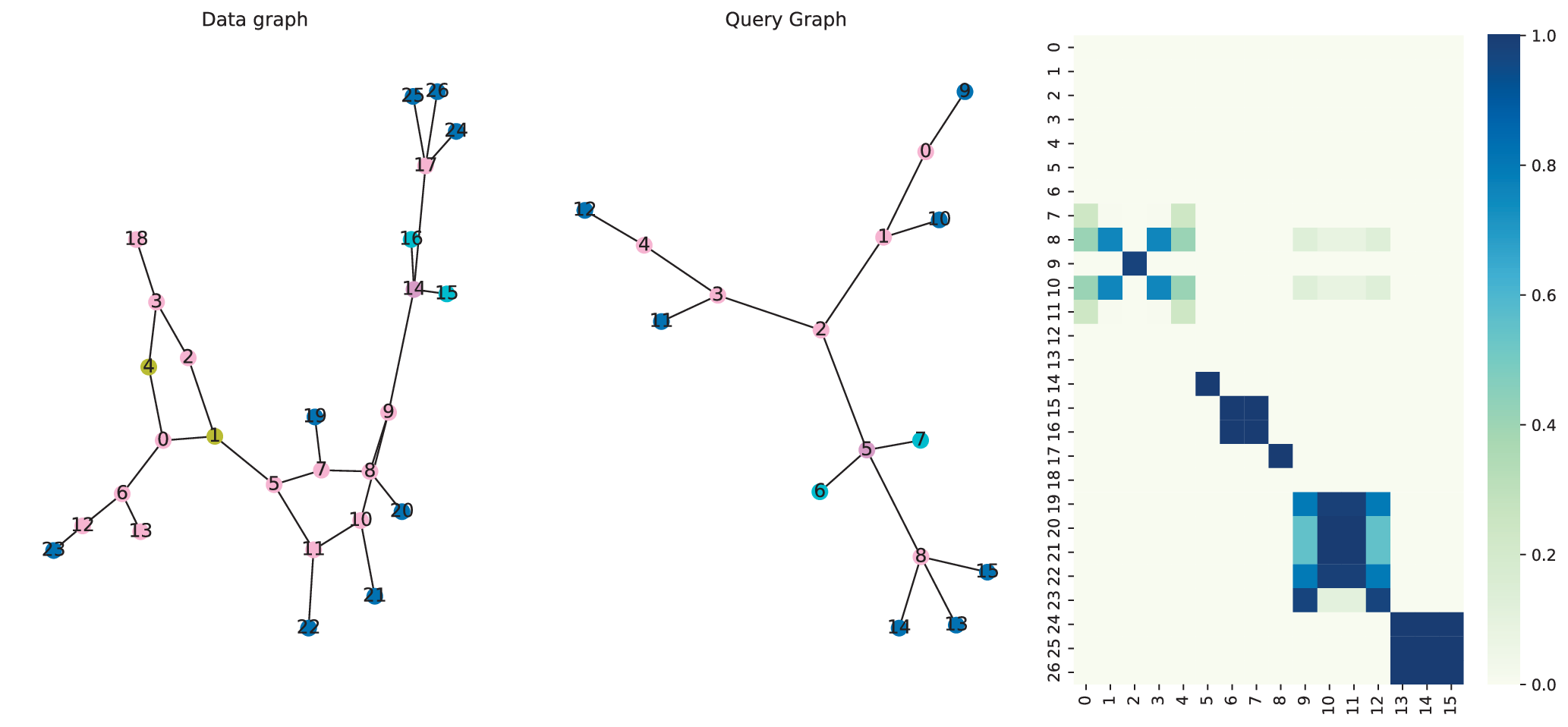}}
   \caption{\footnotesize NC-iso on matched pair.}
   \end{subfigure}
   \begin{subfigure}{0.49\linewidth}{
   \includegraphics[width=1\linewidth]{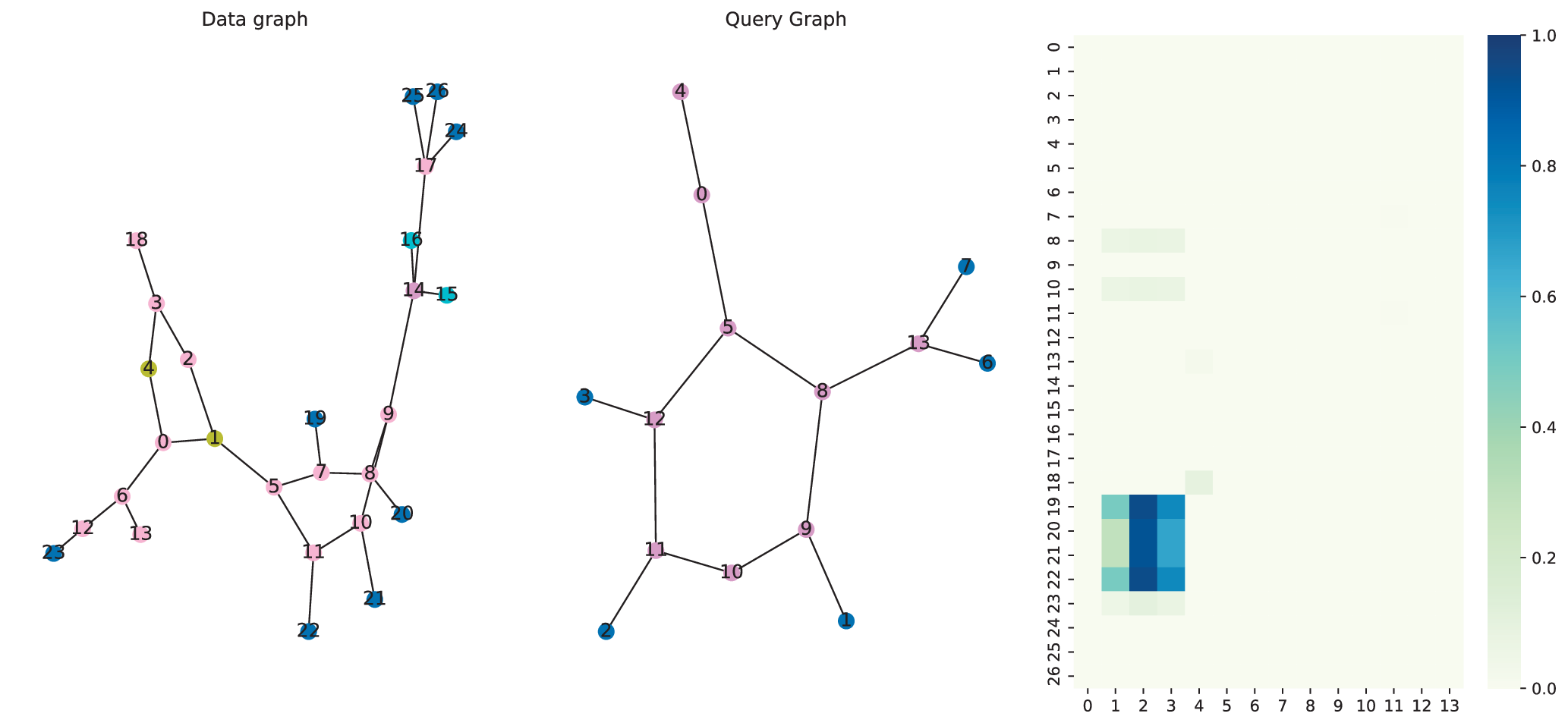}}
   \caption{\footnotesize NC-iso on unmatched pair.}
   \end{subfigure}

   \begin{subfigure}{0.49\linewidth}{
   \includegraphics[width=1\linewidth]{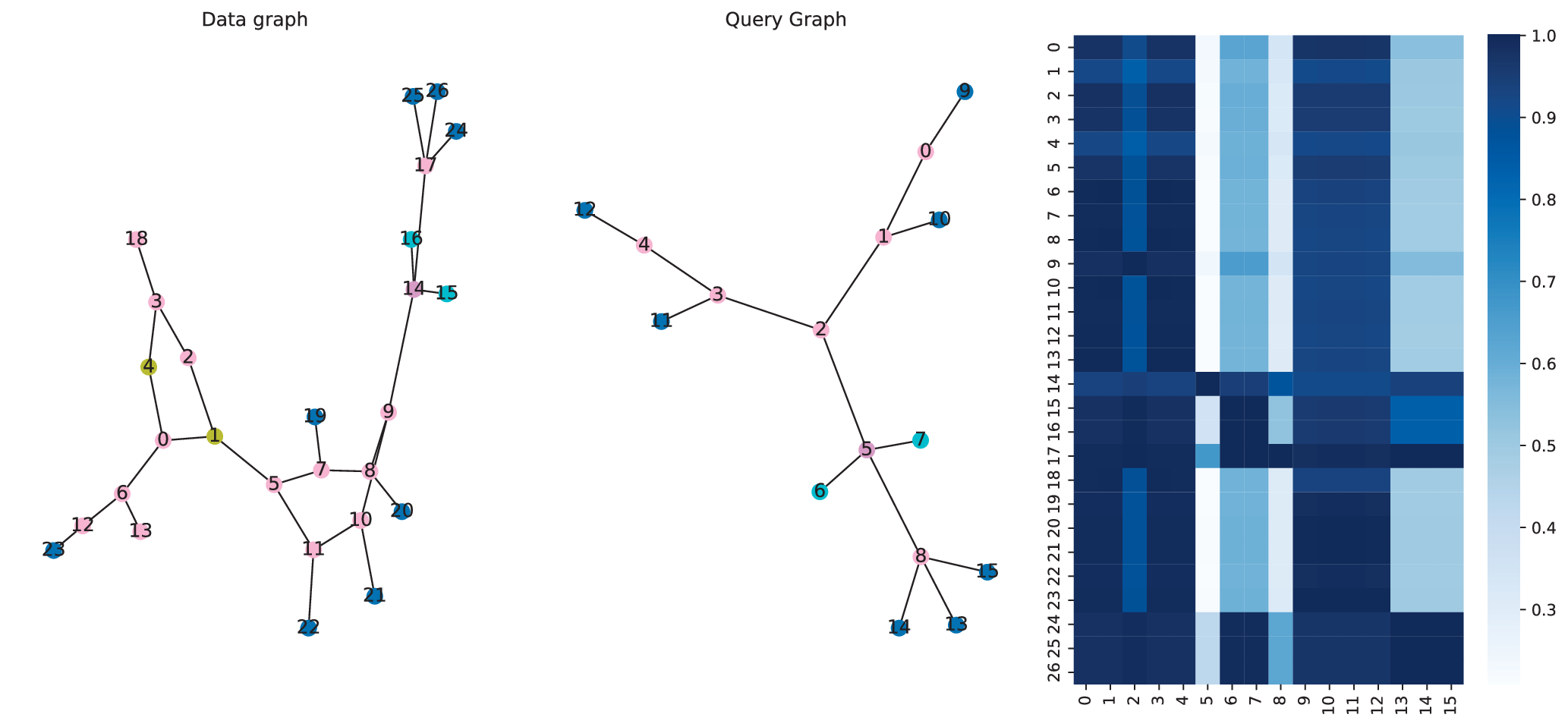}}
   \caption{\footnotesize Greed on matched pair.}
   \end{subfigure}
   \begin{subfigure}{0.49\linewidth}{
   \includegraphics[width=1\linewidth]{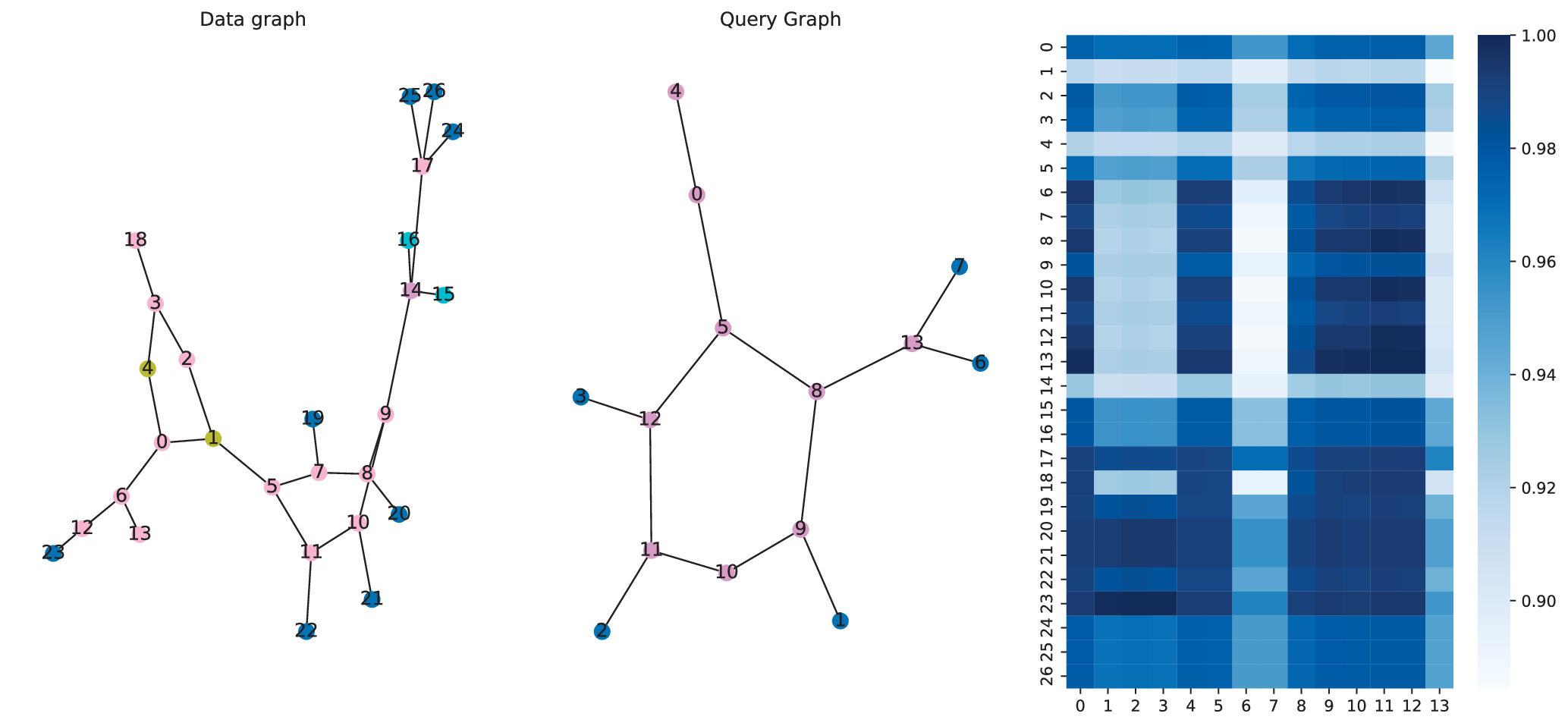}}
   \caption{\footnotesize Greed on matched pair.}
   \end{subfigure}

   \begin{subfigure}{0.49\linewidth}{
   \includegraphics[width=1\linewidth]{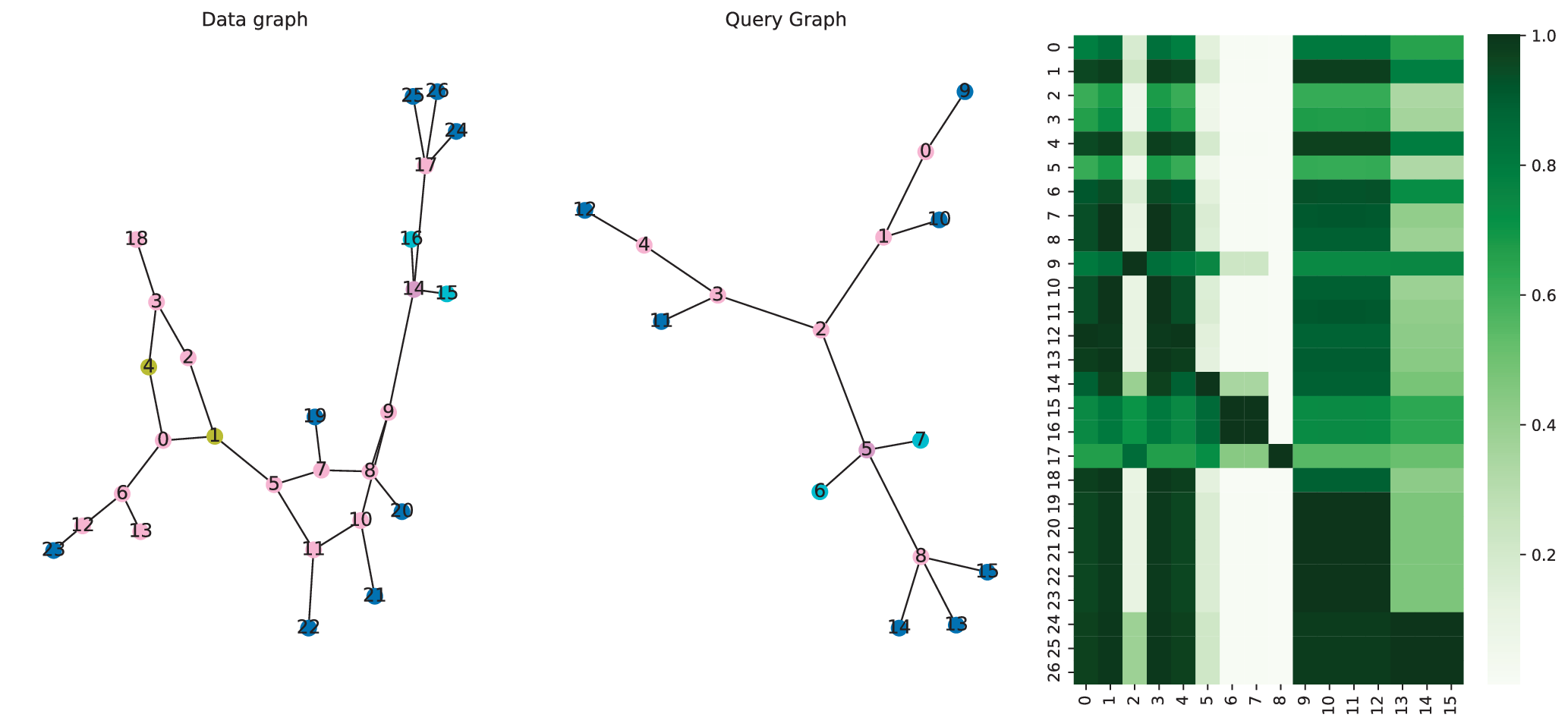}}
   \caption{\footnotesize NeuroMatch on matched pair.}
   \end{subfigure}
   \begin{subfigure}{0.49\linewidth}{
   \includegraphics[width=1\linewidth]{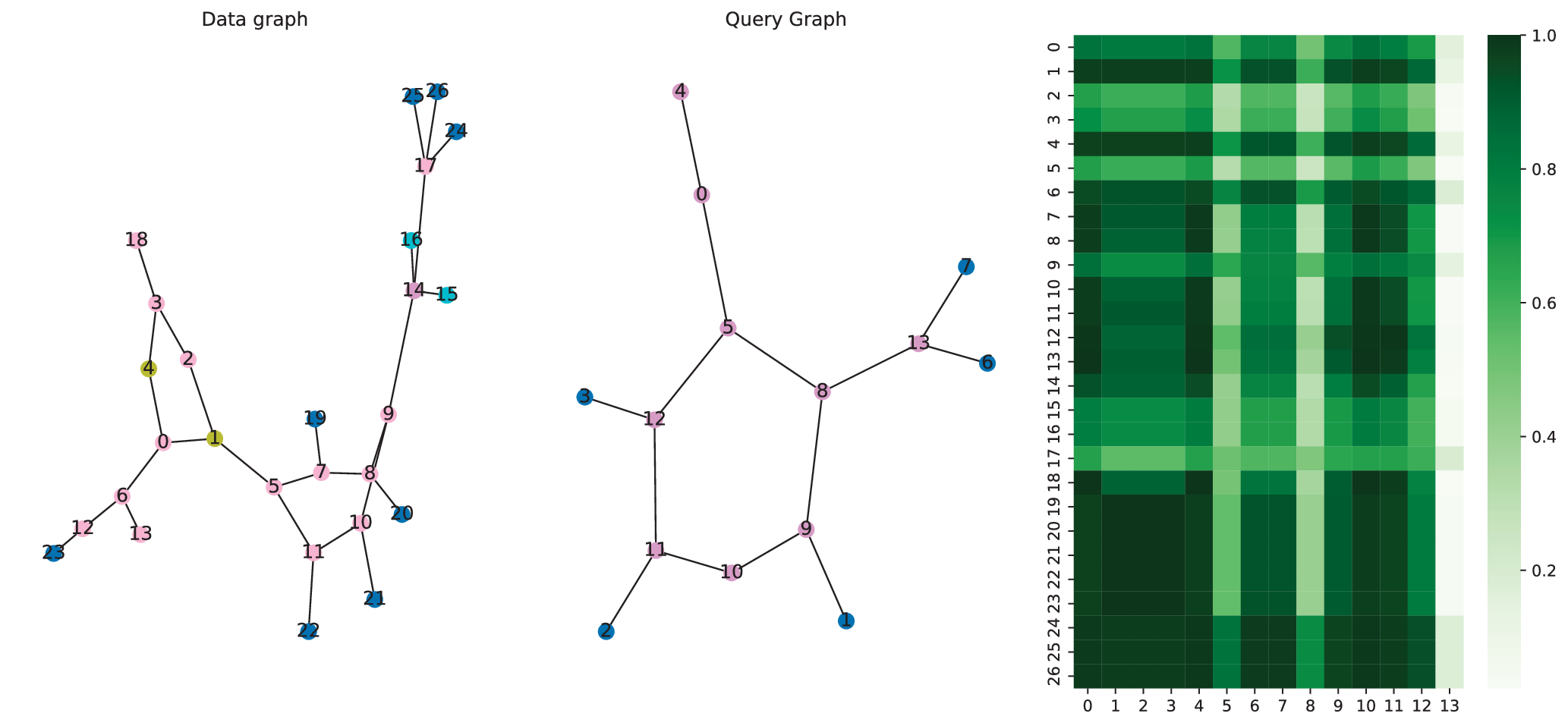}}
   \caption{\footnotesize NeuroMatch on unmatched pair.}
   \end{subfigure}
   \caption{Case study on Cox2 dataset. We present the node pair similarity of graph pairs. The deeper color, the higher similarity score.}\label{fig:ccox2}
   \vspace{-0.5cm}
\end{figure}

\begin{figure}[tb]
   \begin{subfigure}{0.49\linewidth}{
   \includegraphics[width=1\linewidth]{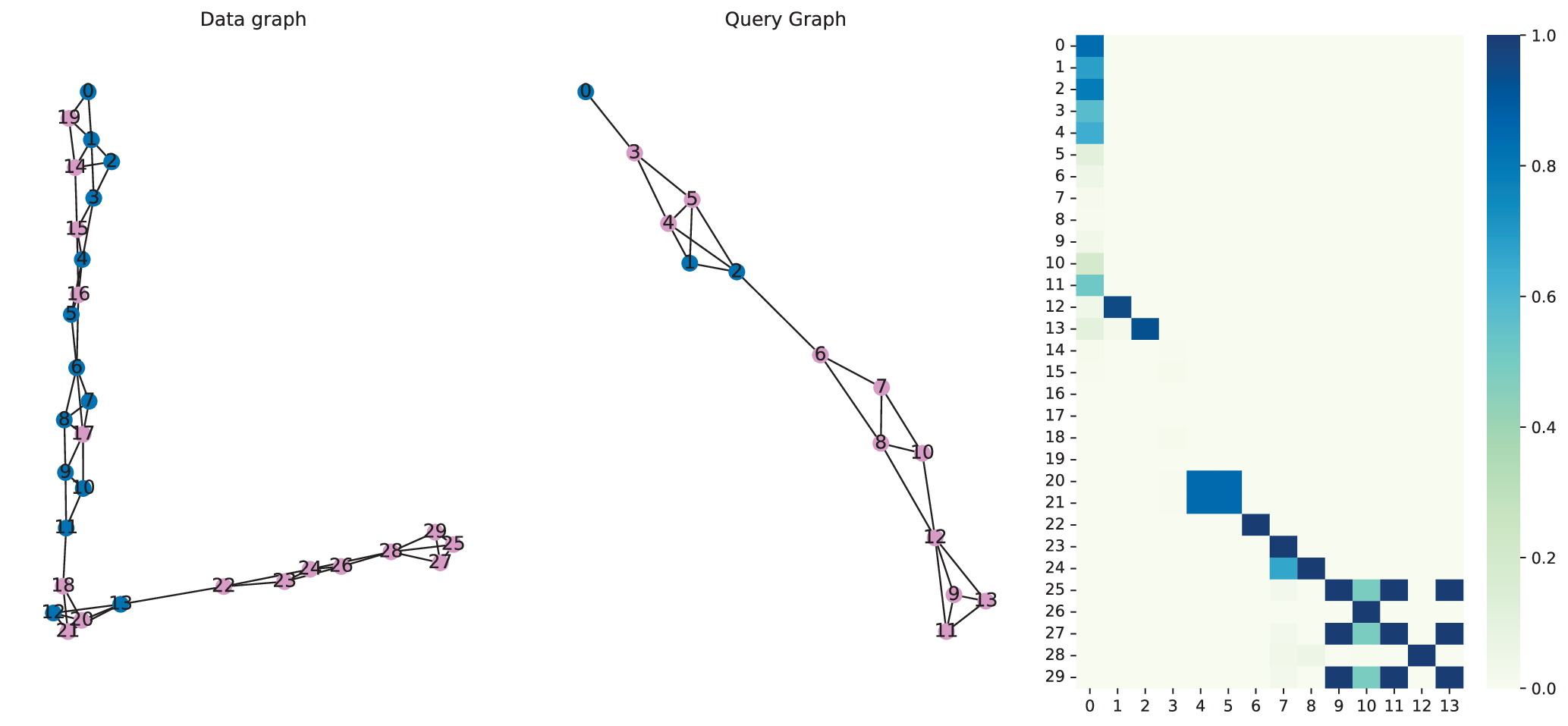}}
   \caption{\footnotesize NC-iso on matched pair.}
   \end{subfigure}
   \begin{subfigure}{0.49\linewidth}{
   \includegraphics[width=1\linewidth]{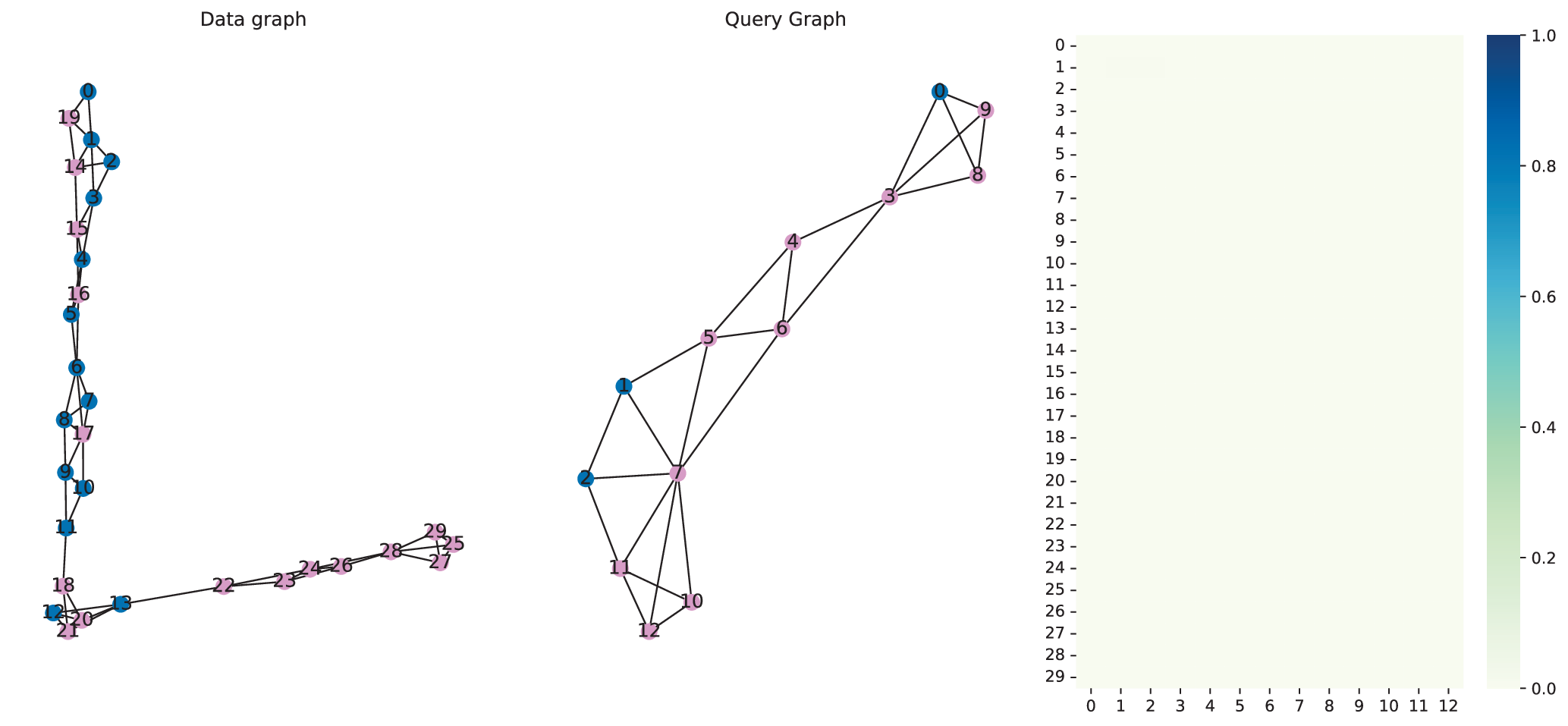}}
   \caption{\footnotesize NC-iso on unmatched pair.}
   \end{subfigure}

   \begin{subfigure}{0.49\linewidth}{
   \includegraphics[width=1\linewidth]{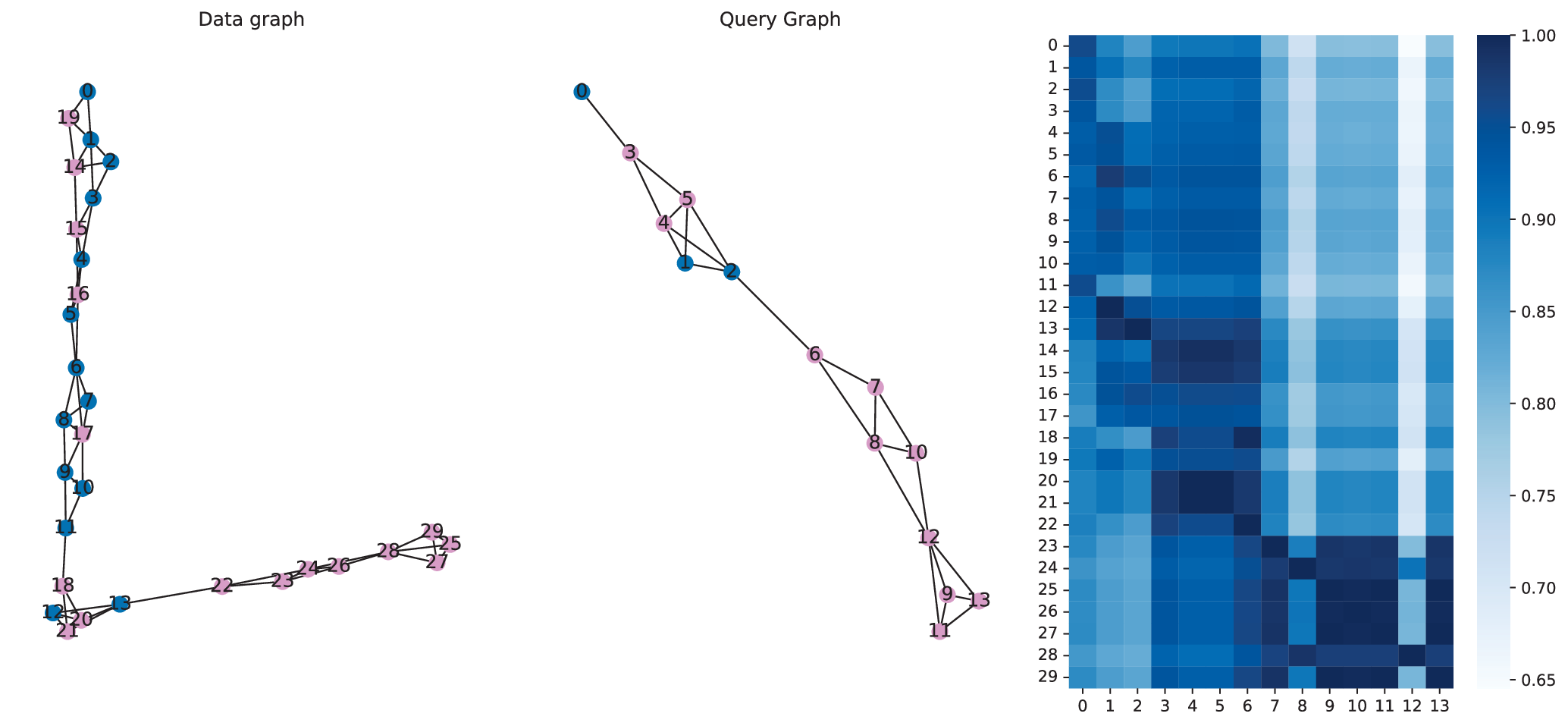}}
   \caption{\footnotesize Greed on matched pair.}
   \end{subfigure}
   \begin{subfigure}{0.49\linewidth}{
   \includegraphics[width=1\linewidth]{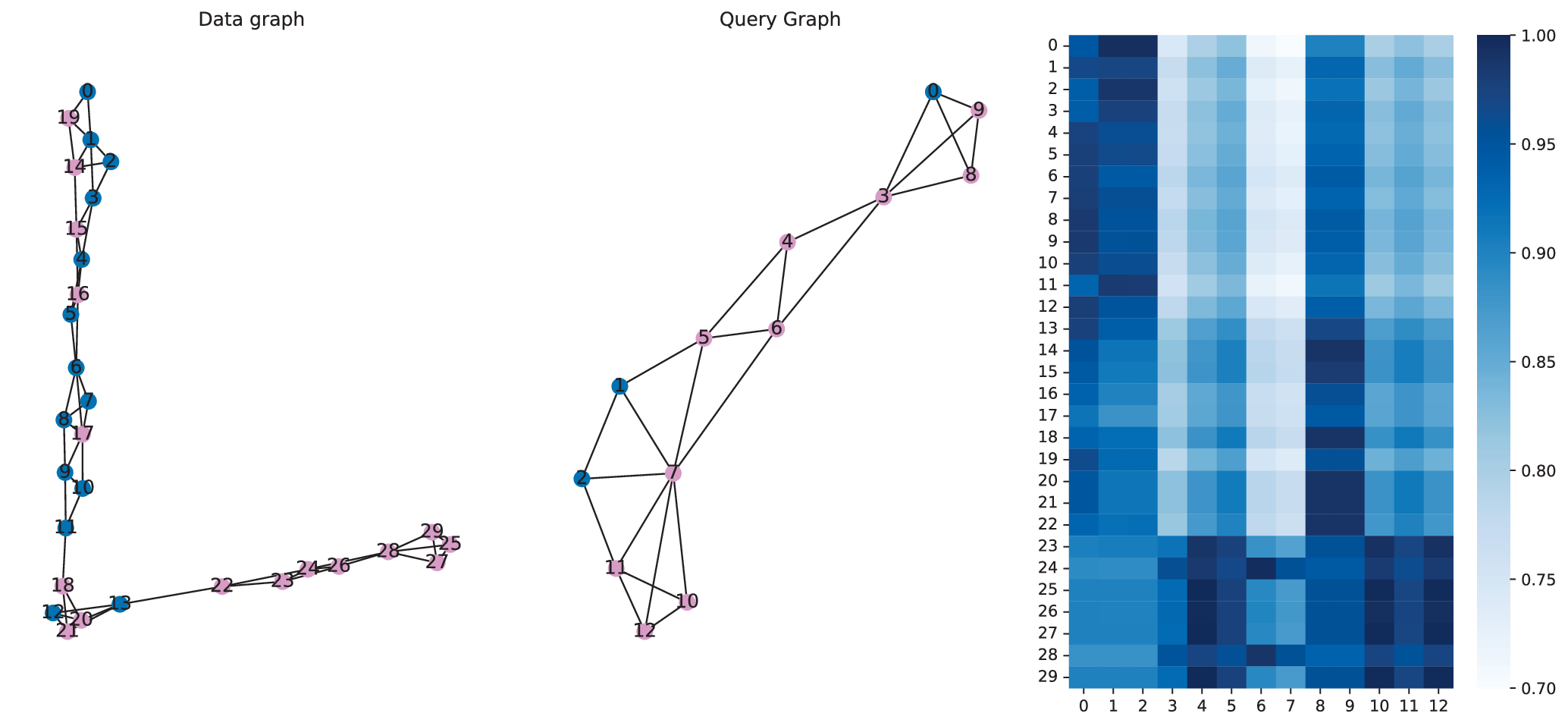}}
   \caption{\footnotesize Greed on matched pair.}
   \end{subfigure}

   \begin{subfigure}{0.49\linewidth}{
   \includegraphics[width=1\linewidth]{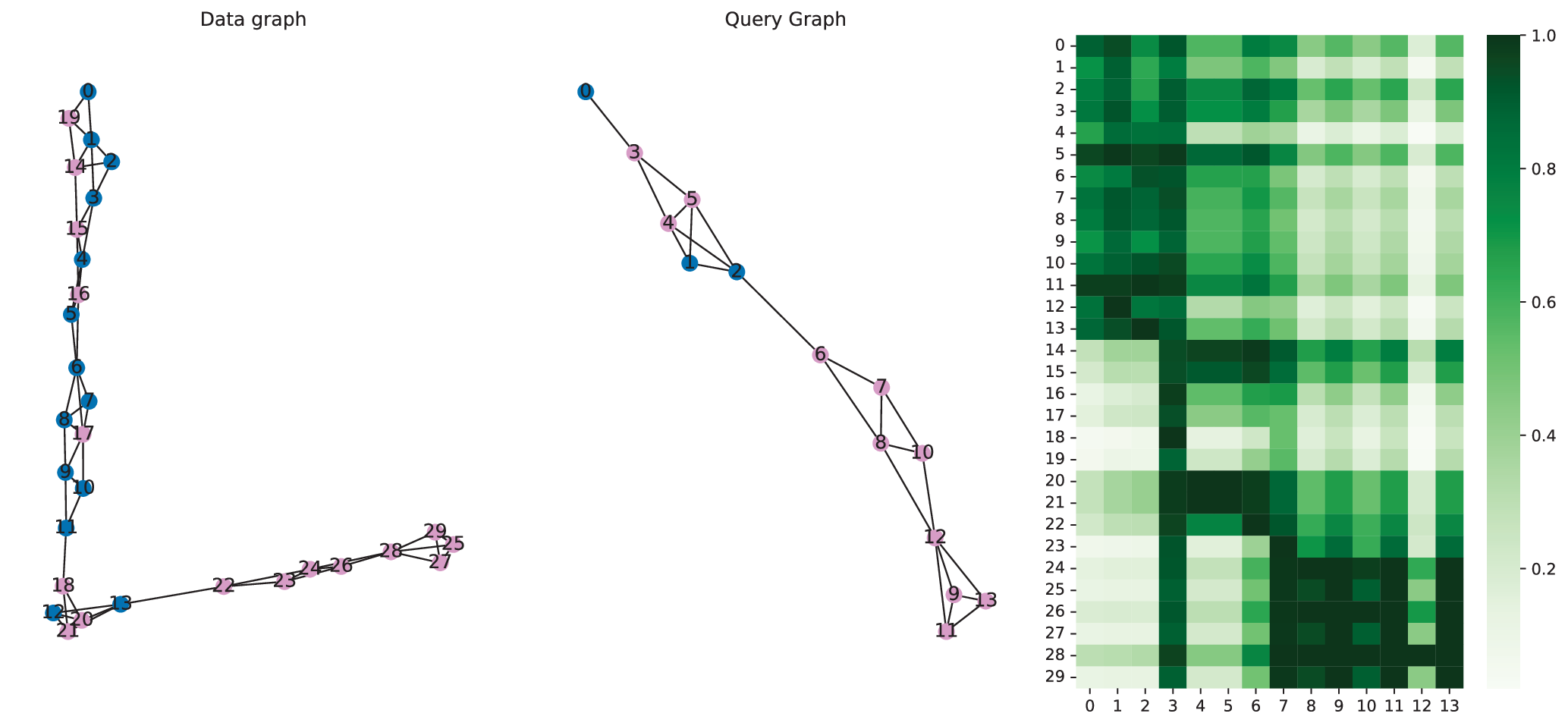}}
   \caption{\footnotesize NeuroMatch on matched pair.}
   \end{subfigure}
   \begin{subfigure}{0.49\linewidth}{
   \includegraphics[width=1\linewidth]{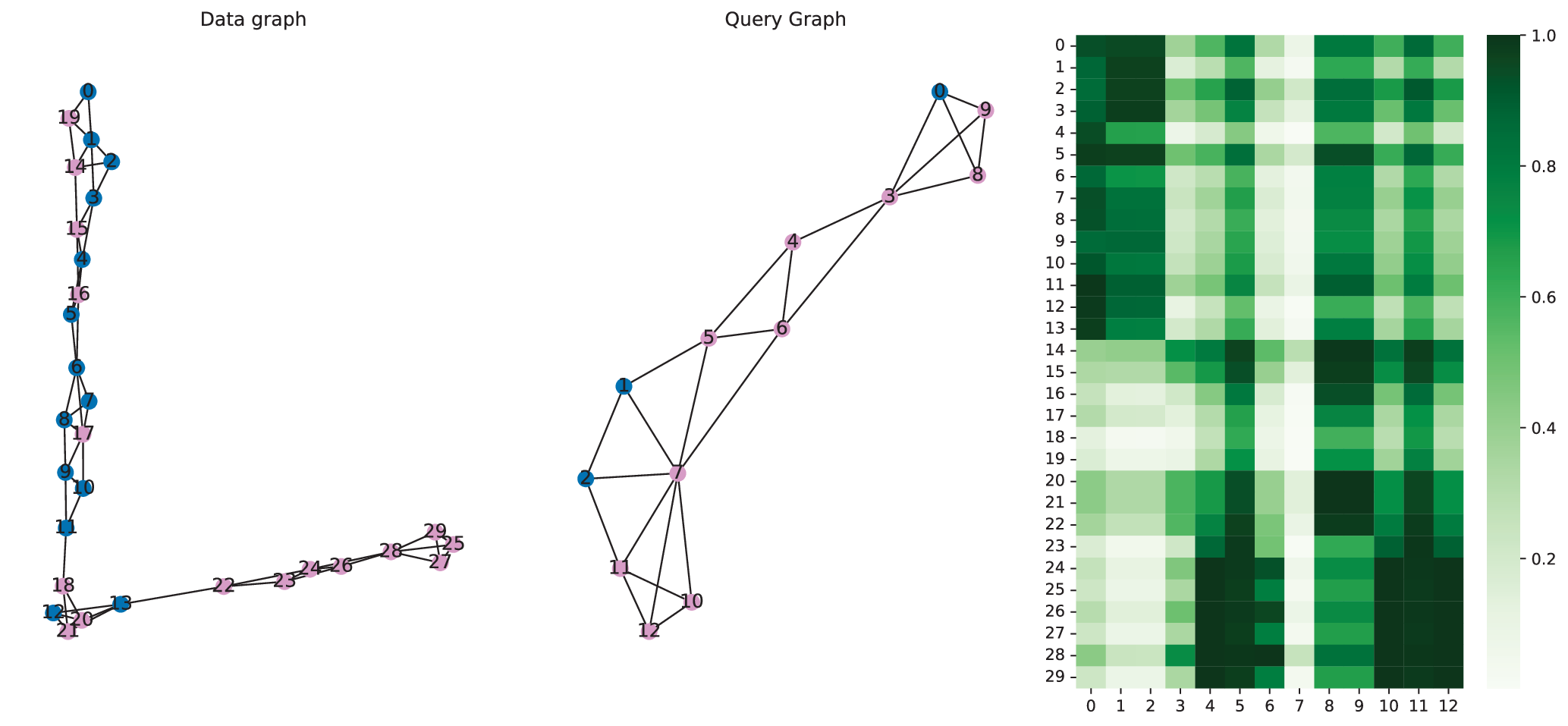}}
   \caption{\footnotesize NeuroMatch on unmatched pair.}
   \end{subfigure}
   \caption{Case study on Enzymes dataset.}\label{fig:enzymes}
\end{figure}

\begin{figure}[tb]
    \vspace{-0.5cm}
    \begin{subfigure}{0.49\linewidth}{
   \includegraphics[width=1\linewidth]{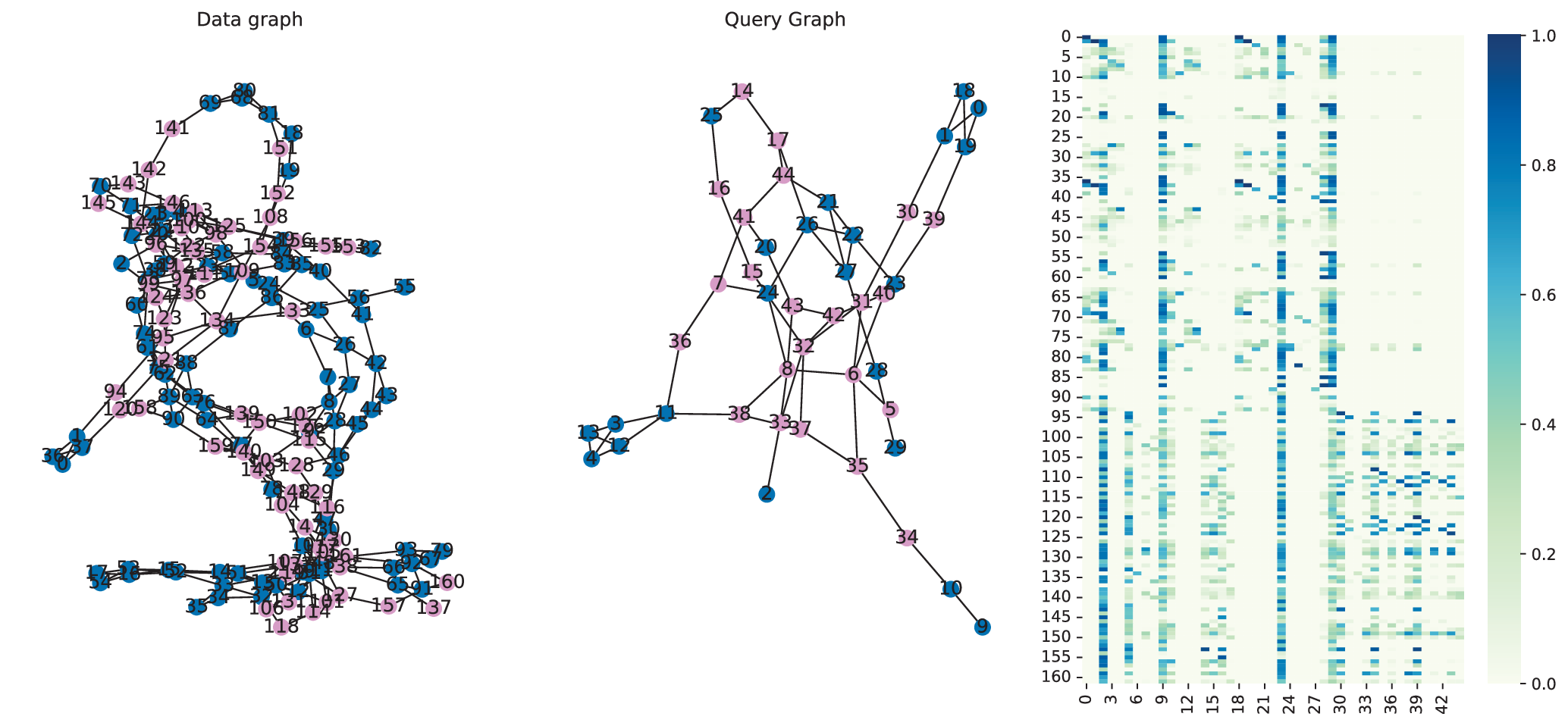}}
   \caption{\footnotesize NC-iso on matched pair.}
   \end{subfigure}
   \begin{subfigure}{0.49\linewidth}{
   \includegraphics[width=1\linewidth]{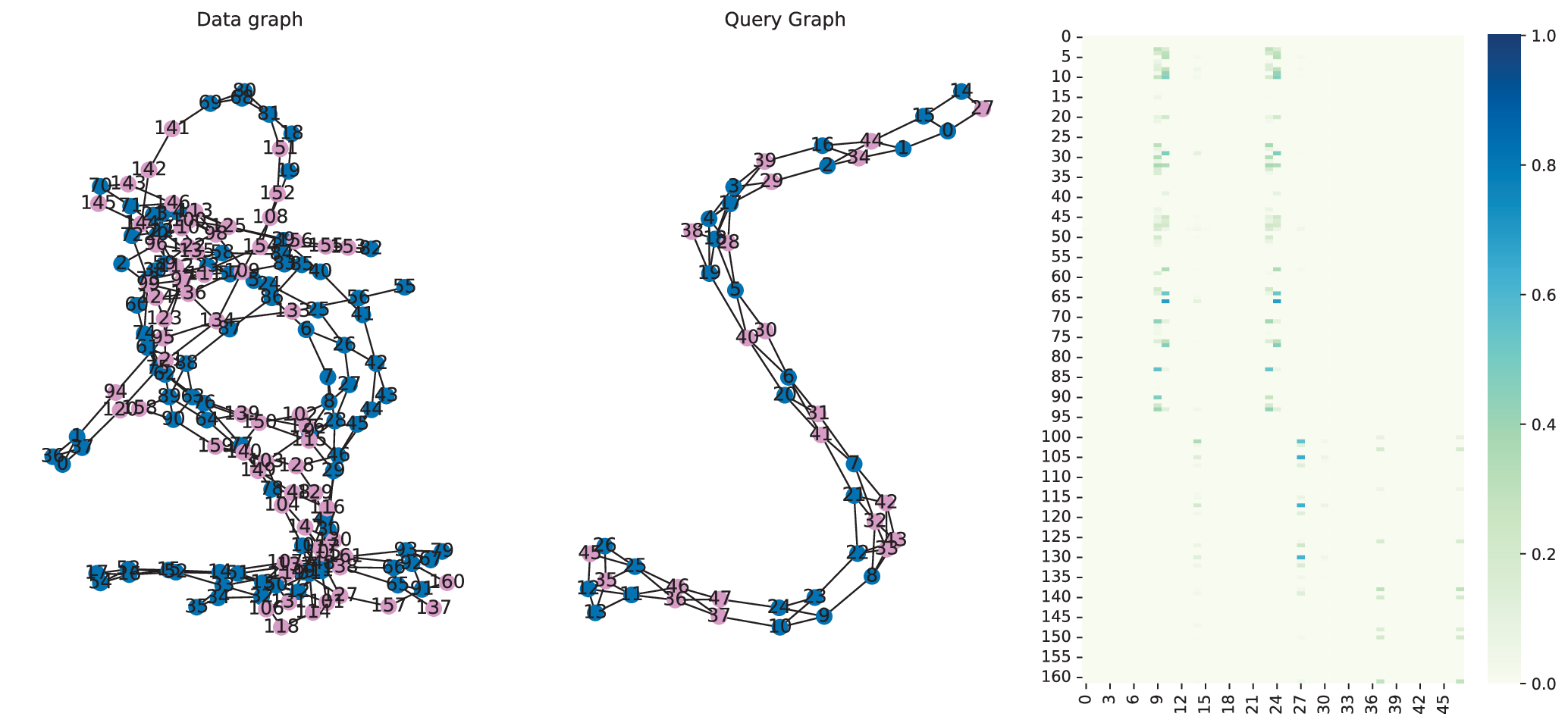}}
    \caption{\footnotesize NC-iso on unmatched pair.}
   \end{subfigure}
   \begin{subfigure}{0.49\linewidth}{
   \includegraphics[width=1\linewidth]{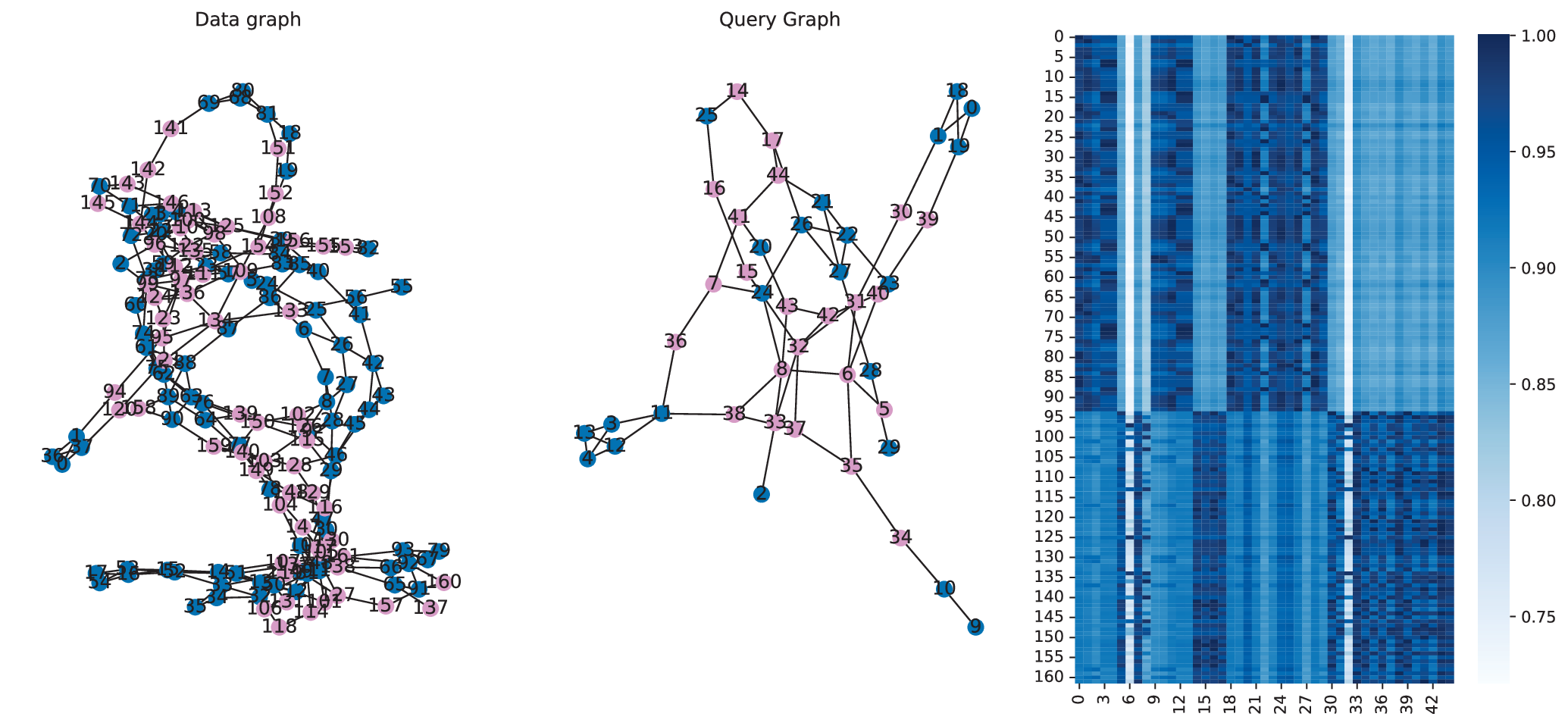}}
   \caption{\footnotesize Greed on matched pair.}
   \end{subfigure}
   \begin{subfigure}{0.49\linewidth}{
   \includegraphics[width=1\linewidth]{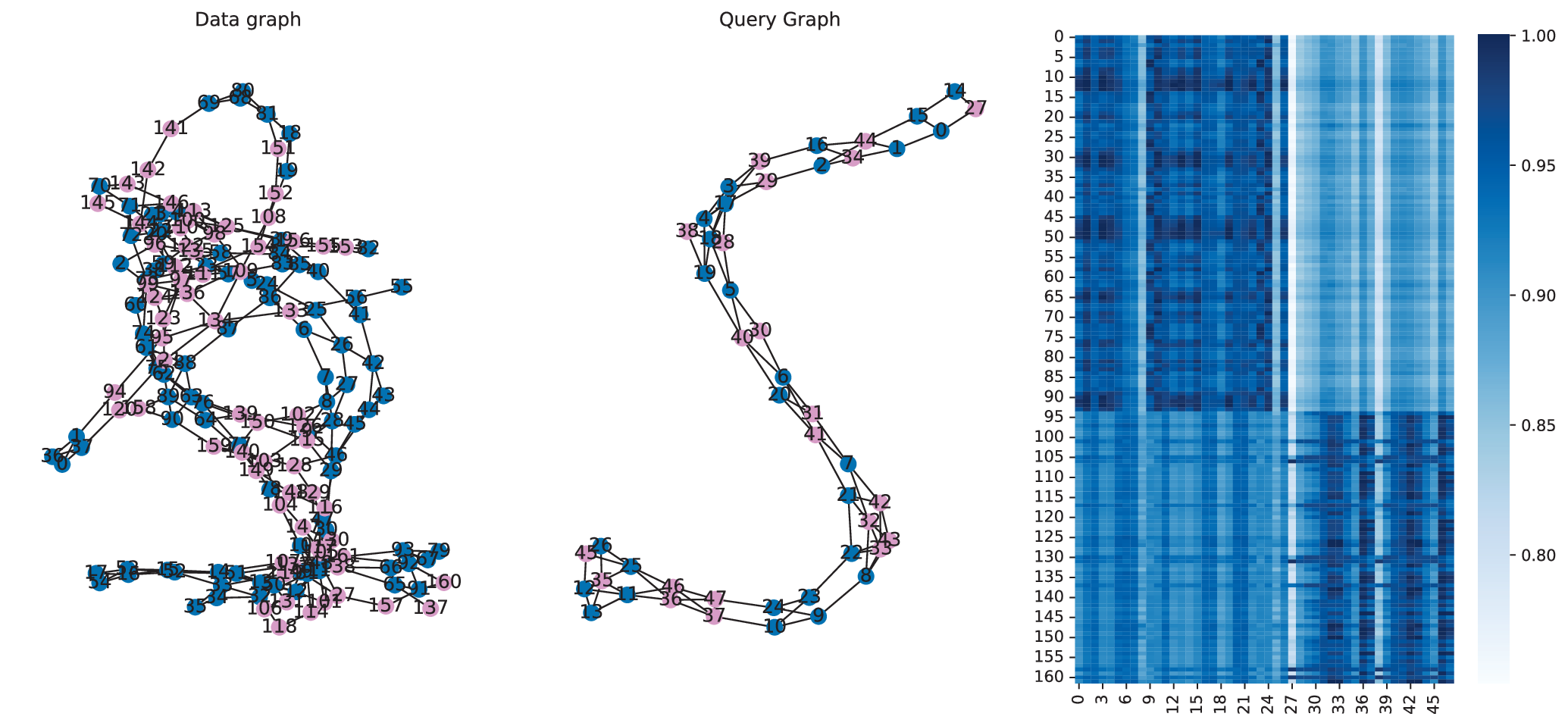}}
   \caption{\footnotesize Greed on unmatched pair.}
   \end{subfigure}

   \begin{subfigure}{0.49\linewidth}{
   \includegraphics[width=1\linewidth]{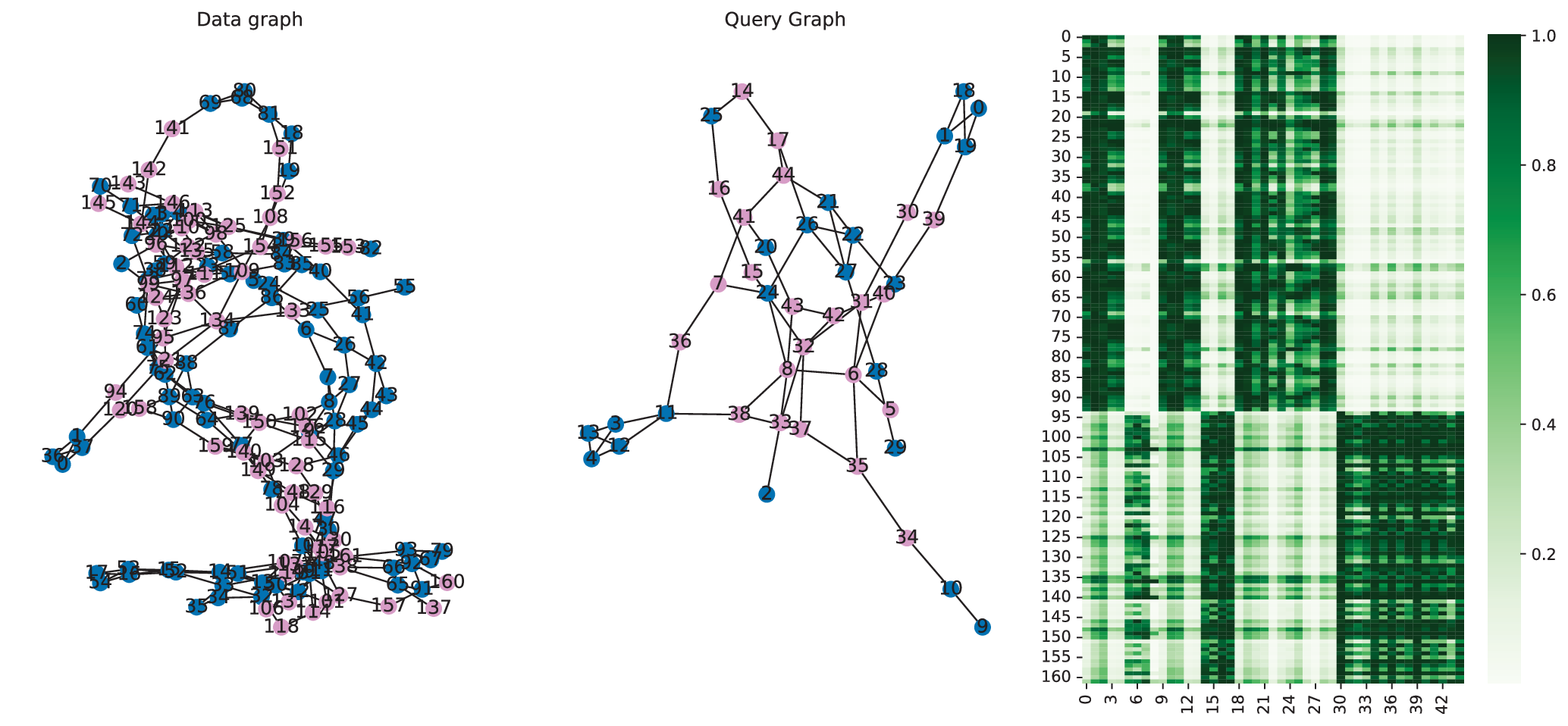}}
   \caption{\footnotesize NeuroMatch on matched pair.}
   \end{subfigure}
   \begin{subfigure}{0.49\linewidth}{
   \includegraphics[width=1\linewidth]{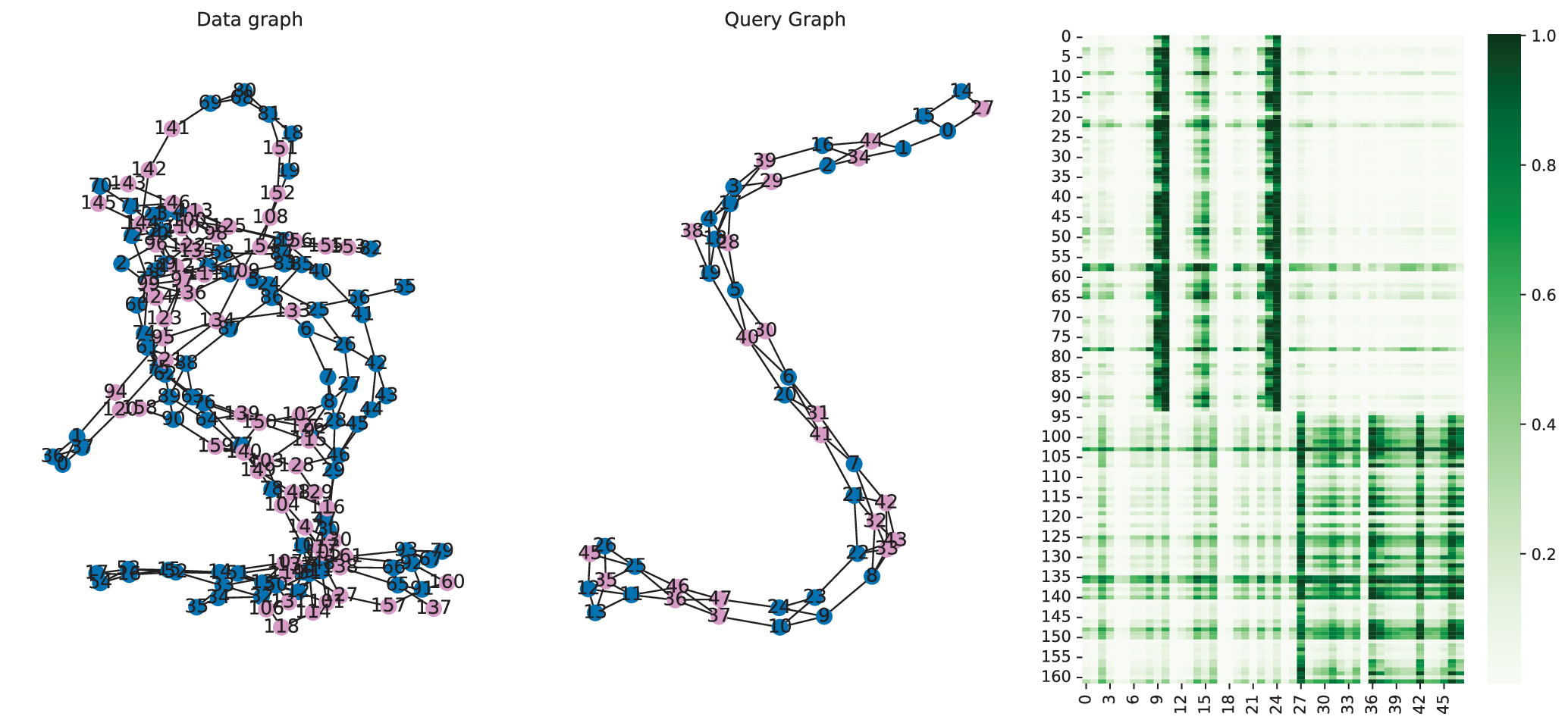}}
   \caption{\footnotesize NeuroMatch on unmatched pair.}
   \end{subfigure}
   \caption{Case study on Proteins dataset.}\label{fig:proteins}
   \vspace{-0.5cm}
\end{figure}

\begin{figure}[tb]
   \begin{subfigure}{0.49\linewidth}{
   \includegraphics[width=1\linewidth]{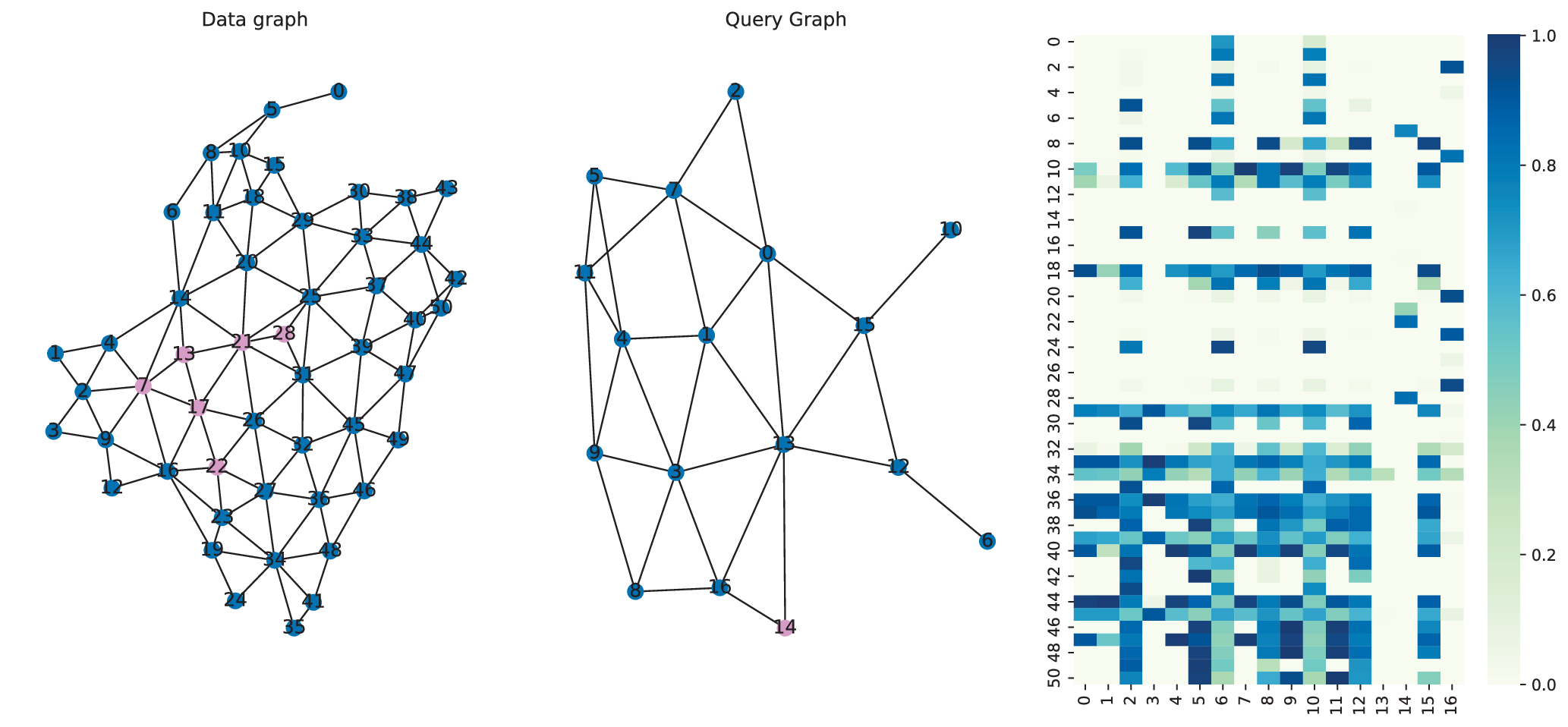}}
   \caption{\footnotesize NC-iso on matched pair.}
   \end{subfigure}
   \begin{subfigure}{0.49\linewidth}{
   \includegraphics[width=1\linewidth]{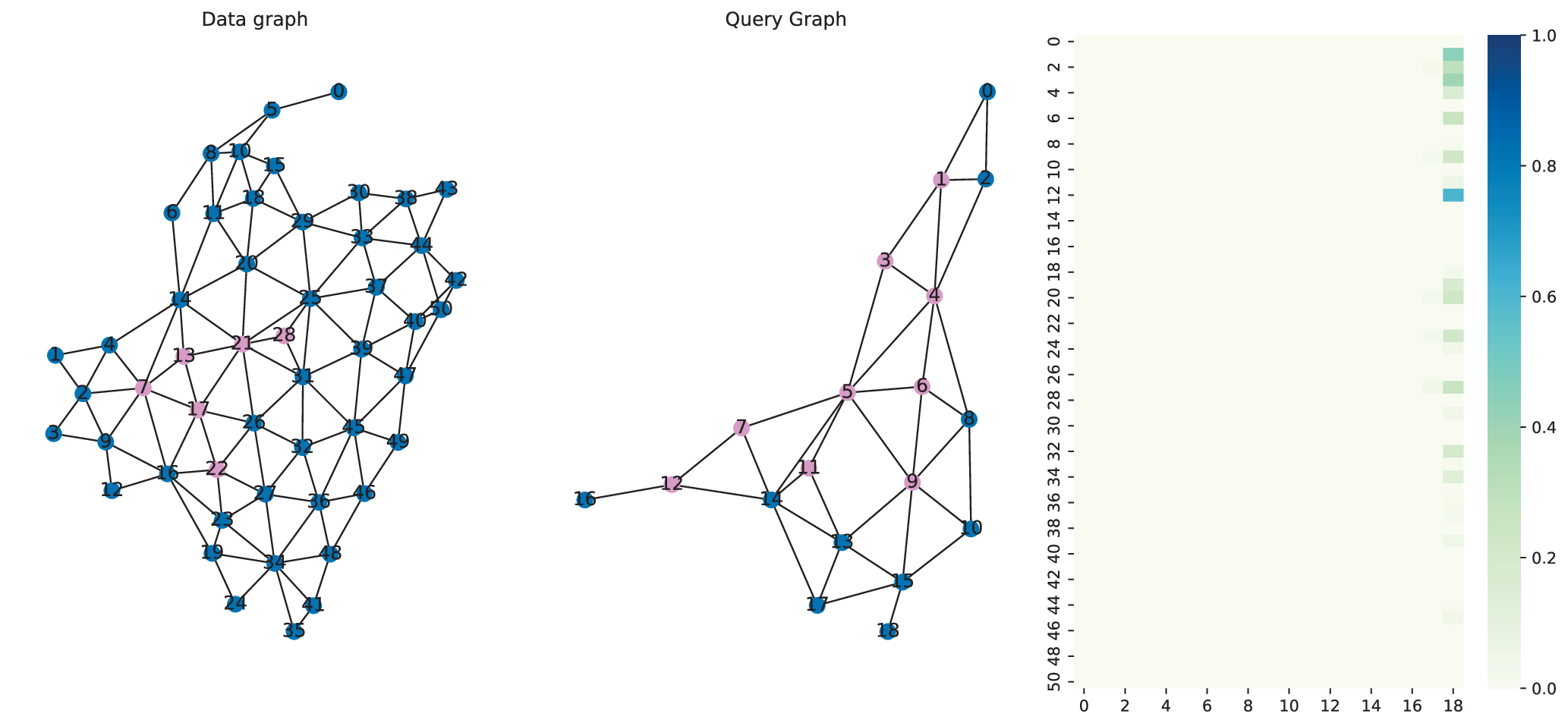}}
   \caption{\footnotesize NC-iso on unmatched pair.}
   \end{subfigure}

   \begin{subfigure}{0.49\linewidth}{
   \includegraphics[width=1\linewidth]{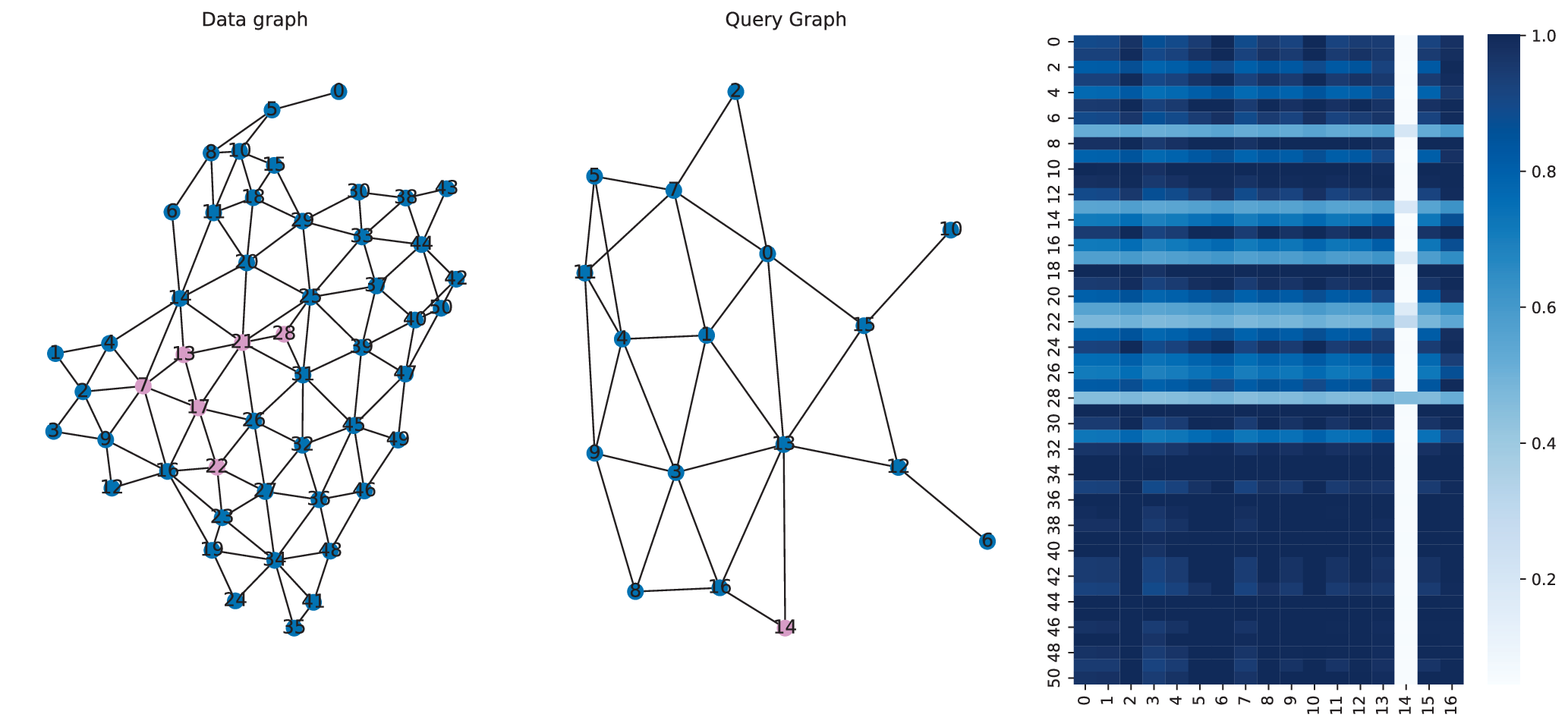}}
   \caption{\footnotesize Greed on matched pair.}
   \end{subfigure}
   \begin{subfigure}{0.49\linewidth}{
   \includegraphics[width=1\linewidth]{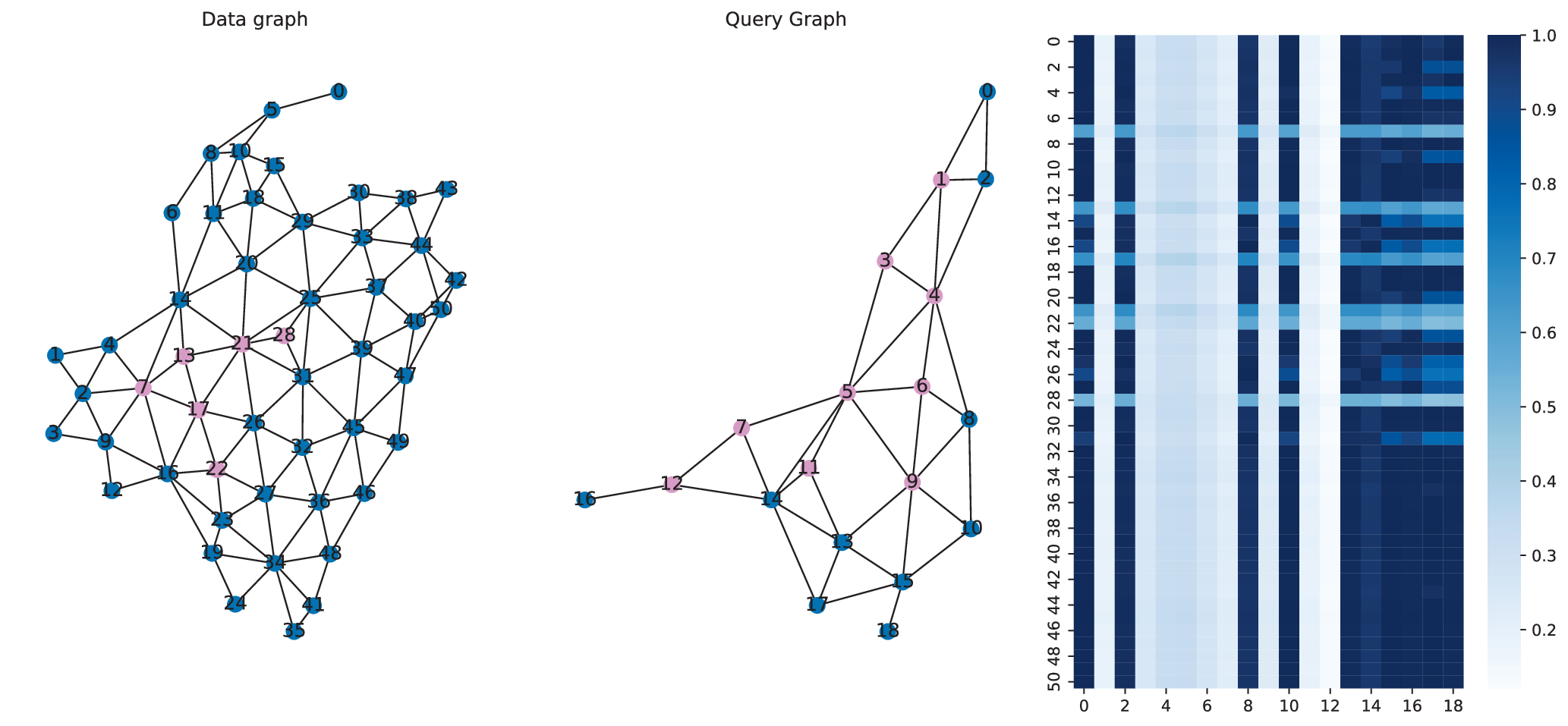}}
   \caption{\footnotesize Greed on matched pair.}
   \end{subfigure}

   \begin{subfigure}{0.49\linewidth}{
   \includegraphics[width=1\linewidth]{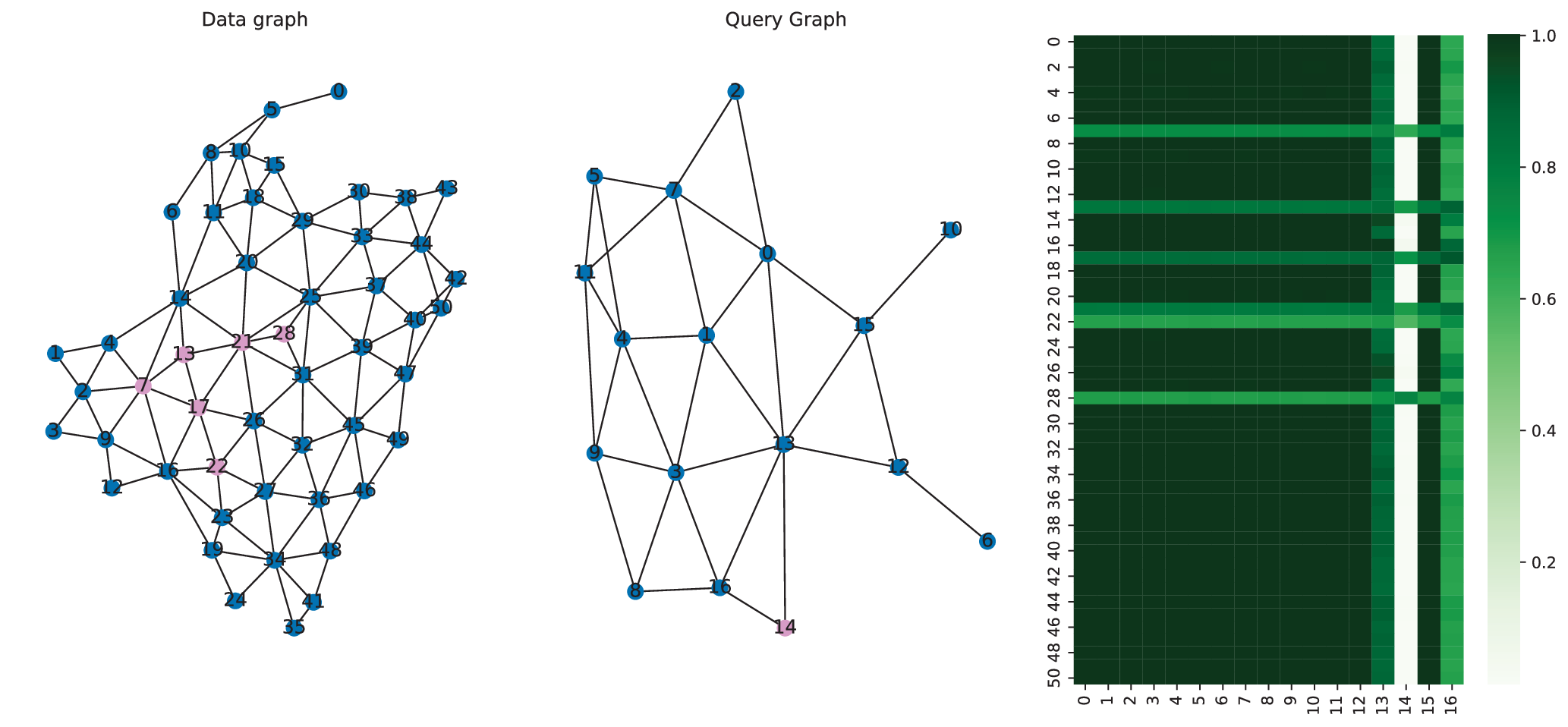}}
   \caption{\footnotesize NeuroMatch on matched pair.}
   \end{subfigure}
   \begin{subfigure}{0.49\linewidth}{
   \includegraphics[width=1\linewidth]{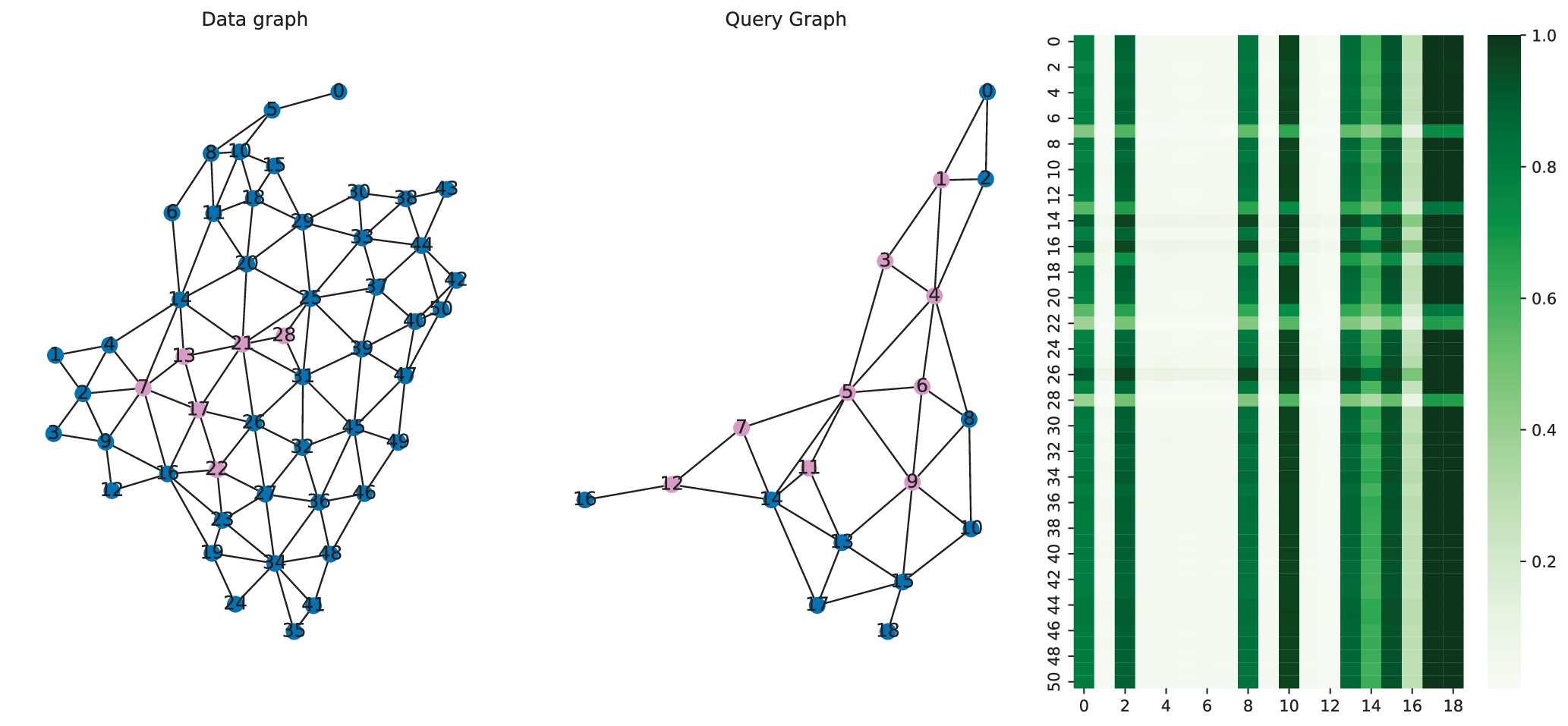}}
   \caption{\footnotesize NeuroMatch on unmatched pair.}
   \end{subfigure}
   \caption{Case study on Msrc\_21 dataset.}\label{fig:cmsrc}
\end{figure}
\subsection{Case Study}
We compare our proposed NC-Iso with NeuroMatch and Greed, as they are graph representation models similar to ours, and they predict subgraph matching with hinge distance measures. The case study results of pairwise node similarity across datasets can be found in Figure \ref{fig:ccox2}, \ref{fig:enzymes}, \ref{fig:proteins} and \ref{fig:cmsrc}.

\section{Conclusion}\label{sec:conclusion}
In this paper, we propose NC-Iso, a simple yet effective architecture for neural subgraph matching. By incorporating a hierarchy-aware Graph Neural Network (GNN) encoder and a novel similarity dominance ratio (SDR)-enhanced measure, NC-Iso reduces the influence of scale differences in the encoding stage and provides a more effective and flexible evaluation of subgraph isomorphism, benefiting applications such as subgraph retrieval. Extensive experiments conducted on diverse datasets have validated the efficacy and efficiency of our proposed approach, surpassing both conventional and neural baseline methods. 

\vspace{-33pt}
\begin{IEEEbiographynophoto}
{Zhouyang Liu} received the BSc (2013-2016) and MSc (2016-2018) degrees in computer science from the Université of Franche-Comté in Besançon, France. She is currently pursuing a PhD in the College of Computer Science and
Technology at the National University of Defense Technology in Changsha, Hunan, China. Her research focuses on graph similarity computation, and graph representation learning.
\end{IEEEbiographynophoto}
\vspace{-1.5cm}
\begin{IEEEbiographynophoto} 
{Ning Liu} received the PhD degree in computer science from the College of Computer Science and Technology, National University of Defense Technology, Changsha, China, in 2023. She is currently an assistant research fellow with the College of Information and Communication, NUDT. Her research interests include graph representation learning, graph OOD generalization, and graph anomaly detection.
\end{IEEEbiographynophoto}
\vspace{-1.5cm}
\begin{IEEEbiographynophoto}
{Yixin Chen} received the B.E. degree in computer science from the National University of Defense Technology in 2009, and the Ph.D. degree from the School of Computer Science at McGill University in 2018. He is currently a research assistant professor at the National University of Defense Technology. His research interests include graph neural networks and multimedia big data systems.
\end{IEEEbiographynophoto}
\vspace{-1.5cm}
\begin{IEEEbiographynophoto}
{Jiezhong He} is currently pursuing a Ph.D. in Computer Science and Technology at the College of Computer, National University of Defense Technology, Changsha, China. His research interests include subgraph matching and graph processing systems.
\end{IEEEbiographynophoto}
\vspace{-2cm}
\begin{IEEEbiographynophoto} 
{Menghan Jia} received the M.Sc. and Ph.D. in computer science from the College of Computer Science, National University of Defense Technology, Changsha, China, in 2019 and 2023, respectively. He is currently a research assistant professor at the National University of Defense Technology. His research interests include graph computing and distributed computing.
\end{IEEEbiographynophoto}
\vspace{-1.5cm}
\begin{IEEEbiographynophoto}
{Dongsheng Li}
received the BSc (with honors) and PhD (with honors) degrees in computer science from the College of Computer, National University
of Defense Technology, Changsha, China, in 1999 and 2005, respectively. He was awarded the prize of National Excellent Doctoral Dissertation of PR China by the Ministry of Education of China in 2008. He is now a full professor at the National Lab for Parallel and Distributed Processing, National University of Defense Technology, China. His research interests include parallel and distributed computing, cloud computing, and large-scale data management.
\end{IEEEbiographynophoto}

\begin{thebibliography}{10}
\bibliographystyle{IEEEtran}
\providecommand{\url}[1]{#1}
\csname url@samestyle\endcsname
\providecommand{\newblock}{\relax}
\providecommand{\bibinfo}[2]{#2}
\providecommand{\BIBentrySTDinterwordspacing}{\spaceskip=0pt\relax}
\providecommand{\BIBentryALTinterwordstretchfactor}{4}
\providecommand{\BIBentryALTinterwordspacing}{\spaceskip=\fontdimen2\font plus
\BIBentryALTinterwordstretchfactor\fontdimen3\font minus
  \fontdimen4\font\relax}
\providecommand{\BIBforeignlanguage}[2]{{%
\expandafter\ifx\csname l@#1\endcsname\relax
\typeout{** WARNING: IEEEtran.bst: No hyphenation pattern has been}%
\typeout{** loaded for the language `#1'. Using the pattern for}%
\typeout{** the default language instead.}%
\else
\language=\csname l@#1\endcsname
\fi
#2}}
\providecommand{\BIBdecl}{\relax}
\BIBdecl

\bibitem{Ullmann}
\BIBentryALTinterwordspacing
J.~R. Ullmann, ``An algorithm for subgraph isomorphism,'' \emph{J. ACM},
  vol.~23, no.~1, p. 31–42, jan 1976. [Online]. Available:
  \url{https://doi.org/10.1145/321921.321925}
\BIBentrySTDinterwordspacing

\bibitem{VF2}
L.~P. Cordella, P.~Foggia, C.~Sansone, and M.~Vento, ``An improved algorithm
  for matching large graphs,'' 2001.

\bibitem{SAGA}
\BIBentryALTinterwordspacing
Y.~Tian, R.~C. Mceachin, C.~Santos, D.~J. States, and J.~M. Patel, ``Saga: A
  subgraph matching tool for biological graphs,'' \emph{Bioinformatics},
  vol.~23, no.~2, p. 232–239, jan 2007. [Online]. Available:
  \url{https://doi.org/10.1093/bioinformatics/btl571}
\BIBentrySTDinterwordspacing

\bibitem{QuickSI}
\BIBentryALTinterwordspacing
H.~Shang, Y.~Zhang, X.~Lin, and J.~X. Yu, ``Taming verification hardness: An
  efficient algorithm for testing subgraph isomorphism,'' \emph{Proc. VLDB
  Endow.}, vol.~1, no.~1, p. 364–375, aug 2008. [Online]. Available:
  \url{https://doi.org/10.14778/1453856.1453899}
\BIBentrySTDinterwordspacing

\bibitem{Turboiso}
W.-S. Han, J.~Lee, and J.-H. Lee, ``Turboiso: towards ultrafast and robust
  subgraph isomorphism search in large graph databases,'' in \emph{ACM SIGMOD
  Conference}, 2013.

\bibitem{RI}
\BIBentryALTinterwordspacing
V.~Bonnici, R.~Giugno, A.~Pulvirenti, D.~E. Shasha, and A.~Ferro, ``A subgraph
  isomorphism algorithm and its application to biochemical data.'' \emph{BMC
  Bioinform.}, vol.~14, no. S-7, p. S13, 2013. [Online]. Available:
  \url{http://dblp.uni-trier.de/db/journals/bmcbi/bmcbi14S.html#BonniciGPSF13}
\BIBentrySTDinterwordspacing

\bibitem{CFL}
\BIBentryALTinterwordspacing
F.~Bi, L.~Chang, X.~Lin, L.~Qin, and W.~Zhang, ``Efficient subgraph matching by
  postponing cartesian products,'' in \emph{Proceedings of the 2016
  International Conference on Management of Data}, ser. SIGMOD '16.\hskip 1em
  plus 0.5em minus 0.4em\relax New York, NY, USA: Association for Computing
  Machinery, 2016, p. 1199–1214. [Online]. Available:
  \url{https://doi.org/10.1145/2882903.2915236}
\BIBentrySTDinterwordspacing

\bibitem{VF3}
V.~Carletti, P.~Foggia, A.~Saggese, and M.~Vento, ``Challenging the time
  complexity of exact subgraph isomorphism for huge and dense graphs with
  vf3,'' \emph{IEEE Transactions on Pattern Analysis and Machine Intelligence},
  vol.~40, no.~4, pp. 804--818, 2018.

\bibitem{VF2++}
B.~Archibald, F.~Dunlop, R.~Hoffmann, C.~McCreesh, P.~Prosser, and J.~Trimble,
  ``Sequential and parallel solution-biased search for subgraph algorithms,''
  in \emph{Integration of AI and OR Techniques in Constraint Programming},
  2019.

\bibitem{DP-iso}
\BIBentryALTinterwordspacing
M.~Han, H.~Kim, G.~Gu, K.~Park, and W.-S. Han, ``Efficient subgraph matching:
  Harmonizing dynamic programming, adaptive matching order, and failing set
  together,'' in \emph{Proceedings of the 2019 International Conference on
  Management of Data}, ser. SIGMOD '19.\hskip 1em plus 0.5em minus 0.4em\relax
  New York, NY, USA: Association for Computing Machinery, 2019, p. 1429–1446.
  [Online]. Available: \url{https://doi.org/10.1145/3299869.3319880}
\BIBentrySTDinterwordspacing

\bibitem{Ying2020NeuralSM}
R.~Ying, Z.~Lou, J.~You, C.~Wen, A.~Canedo, and J.~Leskovec, ``Neural subgraph
  matching,'' \emph{ArXiv}, vol. abs/2007.03092, 2020.

\bibitem{Roy2022InterpretableNS}
I.~Roy, V.~S. Velugoti, S.~Chakrabarti, and A.~De, ``Interpretable neural
  subgraph matching for graph retrieval,'' in \emph{AAAI}, 2022.

\bibitem{liu2023d2match}
\BIBentryALTinterwordspacing
X.~Liu, L.~Zhang, J.~Sun, Y.~Yang, and H.~Yang, ``{D}2{M}atch: Leveraging deep
  learning and degeneracy for subgraph matching,'' in \emph{Proceedings of the
  40th International Conference on Machine Learning}, ser. Proceedings of
  Machine Learning Research, A.~Krause, E.~Brunskill, K.~Cho, B.~Engelhardt,
  S.~Sabato, and J.~Scarlett, Eds., vol. 202.\hskip 1em plus 0.5em minus
  0.4em\relax PMLR, 23--29 Jul 2023, pp. 22\,454--22\,472. [Online]. Available:
  \url{https://proceedings.mlr.press/v202/liu23ba.html}
\BIBentrySTDinterwordspacing

\bibitem{Hjorth2005TJS}
G.~Hjorth, ``T. jech. set theory. the third millennium edition, revised and
  expanded. springer-verlag, berlin, 2003, viii + 769 pp.'' \emph{Bulletin of
  Symbolic Logic}, vol.~11, pp. 243 -- 245, 2005.

\bibitem{rank4}
S.~Lalithsena, P.~Kapanipathi, and A.~Sheth, ``Harnessing relationships for
  domain-specific subgraph extraction: A recommendation use case,'' in
  \emph{2016 IEEE International Conference on Big Data (Big Data)}, 2016, pp.
  706--715.

\bibitem{rank5}
\BIBentryALTinterwordspacing
X.~Yang, D.~Ajwani, W.~Gatterbauer, P.~K. Nicholson, M.~Riedewald, and A.~Sala,
  ``Any-k: Anytime top-k tree pattern retrieval in labeled graphs,'' in
  \emph{Proceedings of the 2018 World Wide Web Conference}, ser. WWW '18.\hskip
  1em plus 0.5em minus 0.4em\relax Republic and Canton of Geneva, CHE:
  International World Wide Web Conferences Steering Committee, 2018, p.
  489–498. [Online]. Available: \url{https://doi.org/10.1145/3178876.3186115}
\BIBentrySTDinterwordspacing

\bibitem{rank3}
\BIBentryALTinterwordspacing
S.~Ranu and A.~K. Singh, ``Indexing and mining topological patterns for drug
  discovery,'' in \emph{Proceedings of the 15th International Conference on
  Extending Database Technology}, ser. EDBT '12.\hskip 1em plus 0.5em minus
  0.4em\relax New York, NY, USA: Association for Computing Machinery, 2012, p.
  562–565. [Online]. Available: \url{https://doi.org/10.1145/2247596.2247666}
\BIBentrySTDinterwordspacing

\bibitem{rank1}
\BIBentryALTinterwordspacing
L.~Zou, L.~Chen, and Y.~Lu, ``Top-k subgraph matching query in a large graph,''
  in \emph{Proceedings of the ACM First Ph.D. Workshop in CIKM}, ser. PIKM
  '07.\hskip 1em plus 0.5em minus 0.4em\relax New York, NY, USA: Association
  for Computing Machinery, 2007, p. 139–146. [Online]. Available:
  \url{https://doi.org/10.1145/1316874.1316897}
\BIBentrySTDinterwordspacing

\bibitem{rank2}
\BIBentryALTinterwordspacing
W.~Fan, X.~Wang, and Y.~Wu, ``Diversified top-k graph pattern matching,''
  \emph{Proc. VLDB Endow.}, vol.~6, no.~13, p. 1510–1521, aug 2013. [Online].
  Available: \url{https://doi.org/10.14778/2536258.2536263}
\BIBentrySTDinterwordspacing

\bibitem{giou}
H.~Rezatofighi, N.~Tsoi, J.~Gwak, A.~Sadeghian, I.~Reid, and S.~Savarese,
  ``Generalized intersection over union: A metric and a loss for bounding box
  regression,'' in \emph{2019 IEEE/CVF Conference on Computer Vision and
  Pattern Recognition (CVPR)}, 2019, pp. 658--666.

\bibitem{Ness}
\BIBentryALTinterwordspacing
A.~Khan, N.~Li, X.~Yan, Z.~Guan, S.~Chakraborty, and S.~Tao, ``Neighborhood
  based fast graph search in large networks,'' in \emph{Proceedings of the 2011
  ACM SIGMOD International Conference on Management of Data}, ser. SIGMOD
  '11.\hskip 1em plus 0.5em minus 0.4em\relax New York, NY, USA: Association
  for Computing Machinery, 2011, p. 901–912. [Online]. Available:
  \url{https://doi.org/10.1145/1989323.1989418}
\BIBentrySTDinterwordspacing

\bibitem{NeMa}
\BIBentryALTinterwordspacing
A.~Khan, Y.~Wu, C.~C. Aggarwal, and X.~Yan, ``Nema: Fast graph search with
  label similarity,'' \emph{Proc. VLDB Endow.}, vol.~6, no.~3, p. 181–192,
  jan 2013. [Online]. Available: \url{https://doi.org/10.14778/2535569.2448952}
\BIBentrySTDinterwordspacing

\bibitem{Vendrov2016OrderEmbeddingsOI}
\BIBentryALTinterwordspacing
I.~Vendrov, R.~Kiros, S.~Fidler, and R.~Urtasun, ``Order-embeddings of images
  and language,'' in \emph{4th International Conference on Learning
  Representations, {ICLR} 2016, San Juan, Puerto Rico, May 2-4, 2016,
  Conference Track Proceedings}, 2016. [Online]. Available:
  \url{http://arxiv.org/abs/1511.06361}
\BIBentrySTDinterwordspacing

\bibitem{Xu2019HowPA}
K.~Xu, W.~Hu, J.~Leskovec, and S.~Jegelka, ``How powerful are graph neural
  networks?'' \emph{ArXiv}, vol. abs/1810.00826, 2019.

\bibitem{Li2019GraphMN}
\BIBentryALTinterwordspacing
Y.~Li, C.~Gu, T.~Dullien, O.~Vinyals, and P.~Kohli, ``Graph matching networks
  for learning the similarity of graph structured objects,'' in
  \emph{Proceedings of the 36th International Conference on Machine Learning},
  ser. Proceedings of Machine Learning Research, K.~Chaudhuri and
  R.~Salakhutdinov, Eds., vol.~97.\hskip 1em plus 0.5em minus 0.4em\relax PMLR,
  09--15 Jun 2019, pp. 3835--3845. [Online]. Available:
  \url{https://proceedings.mlr.press/v97/li19d.html}
\BIBentrySTDinterwordspacing

\bibitem{Bai2019SimGNNAN}
Y.~Bai, H.~Ding, S.~Bian, T.~Chen, Y.~Sun, and W.~Wang, ``Simgnn: A neural
  network approach to fast graph similarity computation,'' \emph{Proceedings of
  the Twelfth ACM International Conference on Web Search and Data Mining},
  2019.

\bibitem{roy2022maximum}
\BIBentryALTinterwordspacing
I.~Roy, S.~Chakrabarti, and A.~De, ``Maximum common subgraph guided graph
  retrieval: Late and early interaction networks,'' in \emph{Advances in Neural
  Information Processing Systems}, A.~H. Oh, A.~Agarwal, D.~Belgrave, and
  K.~Cho, Eds., 2022. [Online]. Available:
  \url{https://openreview.net/forum?id=COAcbu3_k4U}
\BIBentrySTDinterwordspacing

\bibitem{ranjan2022greed}
\BIBentryALTinterwordspacing
R.~Ranjan, S.~Grover, S.~Medya, V.~Chakaravarthy, Y.~Sabharwal, and S.~Ranu,
  ``{GREED}: A neural framework for learning graph distance functions,'' in
  \emph{Advances in Neural Information Processing Systems}, A.~H. Oh,
  A.~Agarwal, D.~Belgrave, and K.~Cho, Eds., 2022. [Online]. Available:
  \url{https://openreview.net/forum?id=3LBxVcnsEkV}
\BIBentrySTDinterwordspacing

\bibitem{eric}
\BIBentryALTinterwordspacing
W.~Zhuo and G.~Tan, ``Efficient graph similarity computation with alignment
  regularization,'' in \emph{Advances in Neural Information Processing
  Systems}, S.~Koyejo, S.~Mohamed, A.~Agarwal, D.~Belgrave, K.~Cho, and A.~Oh,
  Eds., vol.~35.\hskip 1em plus 0.5em minus 0.4em\relax Curran Associates,
  Inc., 2022, pp. 30\,181--30\,193. [Online]. Available:
  \url{https://proceedings.neurips.cc/paper_files/paper/2022/file/c2ce2f2701c10a2b2f2ea0bfa43cfaa3-Paper-Conference.pdf}
\BIBentrySTDinterwordspacing

\bibitem{Morris+2020}
\BIBentryALTinterwordspacing
C.~Morris, N.~M. Kriege, F.~Bause, K.~Kersting, P.~Mutzel, and M.~Neumann,
  ``Tudataset: A collection of benchmark datasets for learning with graphs,''
  in \emph{ICML 2020 Workshop on Graph Representation Learning and Beyond (GRL+
  2020)}, 2020. [Online]. Available: \url{www.graphlearning.io}
\BIBentrySTDinterwordspacing

\bibitem{GraphQL}
\BIBentryALTinterwordspacing
H.~He and A.~K. Singh, ``Graphs-at-a-time: Query language and access methods
  for graph databases,'' in \emph{Proceedings of the 2008 ACM SIGMOD
  International Conference on Management of Data}, ser. SIGMOD '08.\hskip 1em
  plus 0.5em minus 0.4em\relax New York, NY, USA: Association for Computing
  Machinery, 2008, p. 405–418. [Online]. Available:
  \url{https://doi.org/10.1145/1376616.1376660}
\BIBentrySTDinterwordspacing

\bibitem{CECI}
\BIBentryALTinterwordspacing
B.~Bhattarai, H.~Liu, and H.~H. Huang, ``Ceci: Compact embedding cluster index
  for scalable subgraph matching,'' in \emph{Proceedings of the 2019
  International Conference on Management of Data}, ser. SIGMOD '19.\hskip 1em
  plus 0.5em minus 0.4em\relax New York, NY, USA: Association for Computing
  Machinery, 2019, p. 1447–1462. [Online]. Available:
  \url{https://doi.org/10.1145/3299869.3300086}
\BIBentrySTDinterwordspacing

\bibitem{In-Memory}
\BIBentryALTinterwordspacing
S.~Sun and Q.~Luo, ``In-memory subgraph matching: An in-depth study,'' in
  \emph{Proceedings of the 2020 ACM SIGMOD International Conference on
  Management of Data}, ser. SIGMOD '20.\hskip 1em plus 0.5em minus 0.4em\relax
  New York, NY, USA: Association for Computing Machinery, 2020, p. 1083–1098.
  [Online]. Available: \url{https://doi.org/10.1145/3318464.3380581}
\BIBentrySTDinterwordspacing

\bibitem{rapid}
\BIBentryALTinterwordspacing
S.~Sun, X.~Sun, Y.~Che, Q.~Luo, and B.~He, ``Rapidmatch: a holistic approach to
  subgraph query processing,'' \emph{Proc. VLDB Endow.}, vol.~14, no.~2, p.
  176–188, oct 2020. [Online]. Available:
  \url{https://doi.org/10.14778/3425879.3425888}
\BIBentrySTDinterwordspacing

\bibitem{ogb}
\BIBentryALTinterwordspacing
W.~Hu, M.~Fey, M.~Zitnik, Y.~Dong, H.~Ren, B.~Liu, M.~Catasta, and J.~Leskovec,
  ``Open graph benchmark: Datasets for machine learning on graphs,'' in
  \emph{Advances in Neural Information Processing Systems}, H.~Larochelle,
  M.~Ranzato, R.~Hadsell, M.~Balcan, and H.~Lin, Eds., vol.~33.\hskip 1em plus
  0.5em minus 0.4em\relax Curran Associates, Inc., 2020, pp. 22\,118--22\,133.
  [Online]. Available:
  \url{https://proceedings.neurips.cc/paper_files/paper/2020/file/fb60d411a5c5b72b2e7d3527cfc84fd0-Paper.pdf}
\BIBentrySTDinterwordspacing

\end{thebibliography}
\end{document}